\documentclass[10pt,journal,compsoc]{IEEEtran}

\ifCLASSOPTIONcompsoc
  \usepackage[nocompress]{cite}
\else
  \usepackage{cite}
\fi
\ifCLASSINFOpdf
\else
\fi
\usepackage[utf8]{inputenc} 
\usepackage[T1]{fontenc}    
\usepackage{hyperref}       
\usepackage{url}            
\usepackage{booktabs}       
\usepackage{amsfonts}       
\usepackage{nicefrac}       
\usepackage{microtype}      
\usepackage{graphicx}
\usepackage{wrapfig}
\usepackage{amsmath}
\usepackage{algorithm,algpseudocode}
\usepackage{multirow}
\usepackage{wrapfig}
\usepackage{pgf, tikz}
\usepackage{subfig}
\usepackage{flushend}
\begin{document}
%
\title{Self-reinforcing Unsupervised Matching}
%
%
%
%

\author{Jiang~Lu,~
        Lei~Li,~
        and~Changshui~Zhang,~\IEEEmembership{Fellow,~IEEE}
\IEEEcompsocitemizethanks{\IEEEcompsocthanksitem J. Lu, L. Li and C. Zhang are with the Institute for Artificial Intelligence, Tsinghua University (THUAI), Beijing 100084, China, and also with the State Key Laboratory of Intelligence Technologies and Systems, Beijing National Research Center for Information Science and Technologies (BNRist) , Department of Automation, Tsinghua University, Beijing 100084, China (e-mail: lu-j13@mails.tsinghua.edu.cn; lilei17@mails.tsinghua.edu.cn; zcs@mail.tsinghua.edu.cn).
\IEEEcompsocthanksitem Corresponding Author: J. Lu and C. Zhang.

}
}

\IEEEtitleabstractindextext{%
\begin{abstract}
Remarkable gains in deep learning usually rely on tremendous supervised data. Ensuring the modality diversity for one object in training set is critical for the generalization of cutting-edge deep models, but it burdens human with heavy manual labor on data collection and annotation. In addition, some rare or unexpected modalities are new for the current model, causing reduced performance under such emerging modalities.
Inspired by the achievements in speech recognition, psychology and behavioristics, we present a practical solution, \emph{self-reinforcing unsupervised matching} (SUM), to annotate the images with 2D structure-preserving property in an emerging modality by cross-modality matching. This approach requires no any supervision in emerging modality and only one template in seen modality, providing a possible route towards continual learning.
\end{abstract}

}

\maketitle

\IEEEdisplaynontitleabstractindextext

%
\IEEEpeerreviewmaketitle

\IEEEraisesectionheading{\section{Introduction}\label{sec:introduction}}
\IEEEPARstart{D}{eep} learning~\cite{lecun2015deep} is a data-hungry technology, which has scored great achievements in many research fields~\cite{krizhevsky2012imagenet,esteva2017dermatologist,gebru2017using,ghosal2018explainable,norouzzadeh2018automatically}. Congener objects appear in visual images with multifarious modalities, and a wide consensus is that preserving the modality's variety of objects aids in generalizing the deep model~\cite{jiang2014self,li2016fast,wang2018low,cubuk2018autoaugment}. However, collecting and annotating data for one object with various modalities is expensive, time consuming and laborious. In addition, due to the inevitable rarity and suddenness of objects in reality (e.g., a novel Chinese character font designed by calligraphers or a shipment of traffic signs erected on a new outdoor site), some rare or unexpected modalities are new for the current model, causing compromised performance on such emerging modalities~\cite{torralba2011unbiased,gopalan2011domain}. Exploiting one algorithm capable of autonomously annotating the data in an emerging modality can reduce the tedious manual workload and is meaningful for continual learning~\cite{chen2016lifelong}.
A practical convention of coping with the emerging modality is to leverage expertise to manually annotate the acquired emerging data and then configure a finetuning procedure to strengthen the model generalization capability. Nonetheless, can this semiautonomous or human-in-the-loop fashion be fulfilled by artificial intelligent (AI) agents? Critically, it is a promising motivation for continual learning to enable agents to autonomously annotate emerging data. 

Since the shared label space across all modalities, the autonomous annotation for data in an emerging modality is reasonably treated as a cross-modality data matching issue~\cite{shrivastava2011data} or a data classification issue in a novel domain~\cite{gopalan2011domain}. In contrast, the task we focus on presents two intractable challenges. (1) \emph{Data in the emerging modality are fully unsupervised}, corresponding to the desired regime of autonomous intelligence wherein no manual efforts are needed for data annotation. This challenge makes many methods inapplicable, like those based on metric learning~\cite{xing2003distance,weinberger2009distance} or style transfer mapping~\cite{zhang2013writer}, because the explicit cross-modality correspondence is unknown beforehand. (2) \emph{Data in the seen modality are scarce despite supervision (the extreme case is only one sample per object)}, consistent with the AI destination that agents start from scratch using few supervised data and then incrementally learn with data growth. In this context, the technologies for domain adaptation~\cite{gopalan2011domain,tzeng2014deep} tend to deteriorate because the exiguous supervised data struggle to support the acquisition of adequate transferrable knowledge. Without loss of generality, the task discussed here is distilled into the following concise but difficult setting that combines the two challenges: Given one seen modality with only one template per object class and one emerging modality composed of fully unsupervised data, agents are required to reveal the cross-modality data correspondence.

In this work, we propose \textbf{S}elf-reinforcing \textbf{U}nsupervised \textbf{M}atching (SUM) framework to annotate data autonomously by exploring the cross-modality correspondence on both the local feature and the whole image levels. SUM is amenable for visual objects possessing the 2D structure-preserving property, such as Chinese characters and traffic signs. The effectiveness of SUM is demonstrated on two types of computer visual tasks following the aforementioned difficult setting: Chinese character matching by one template and traffic sign matching by one template.

\section{Theory and Methods}
Our inspiration comes from three aspects. The fisrt is the dynamic time warping (DTW) in speech recognition~\cite{sakoe1978dynamic}, which aims at finding an optimal alignment between two given time-dependent sequences under certain restrictions. Accordingly, we propose \emph{dynamic position warping} (DPW) to align two 2D matrix data in an order-preserving fashion. The second is the similarity-association mechanism in psychology~\cite{wundt2014lectures,Rescorla1977Stimulus}, which refers to the transfer process of human thinking from one idea to another maintaining a certain similarity with the former in appearance or essence. Inspiringly, we devise a \emph{local feature adapter} (LoFA) to transfer the local representation between cross-modality objects. The third is the self-reinforcing mechanism in behavioristics~\cite{bandura1974behavior,aknin2012happiness}, which indicates the positive-feedback process that one behavior serves to strengthen itself. We hereby develop the self-reinforcing learning mechanism for both local feature and whole image matching, enabling the optimization of LoFA and alleviating the dilemma caused by a fully unsupervised emerging modality.

\subsection{Problem Description} 
We apply SUM framework to 2D visual images. According to the difficult setting, we denote the image set of the seen modality as $\{\mathbf{x}^s_i\}_{i=1}^N$ and that of the emerging modality as $\{\mathbf{x}^e_i\}_{i=1}^N$, where $\mathbf{x}^s_i$, $\mathbf{x}^e_i\in \mathbb{R}^{h_I\times w_I \times c_I}$ are RGB images with $h_I$ rows, $w_I$ columns and $c_I$ channels. Notably, $\{\mathbf{x}^s_i\}_{i=1}^N$ and $\{\mathbf{x}^e_i\}_{i=1}^N$ share the same label space, and in $\{\mathbf{x}^s_i\}_{i=1}^N$, only one sample per object class is given (i.e., $N$ object classes in total). We know the class label of each $\mathbf{x}^s_i$ but all $\mathbf{x}^e_i$ are unsupervised. The goal is to find the correct cross-modality image correspondence between $\{\mathbf{x}^s_i\}_{i=1}^N$ and $\{\mathbf{x}^e_i\}_{i=1}^N$.

\subsection{Modeling on Level of Cognitive Psychology}
\begin{figure}[t!]
\centering
\includegraphics[width=1\linewidth]{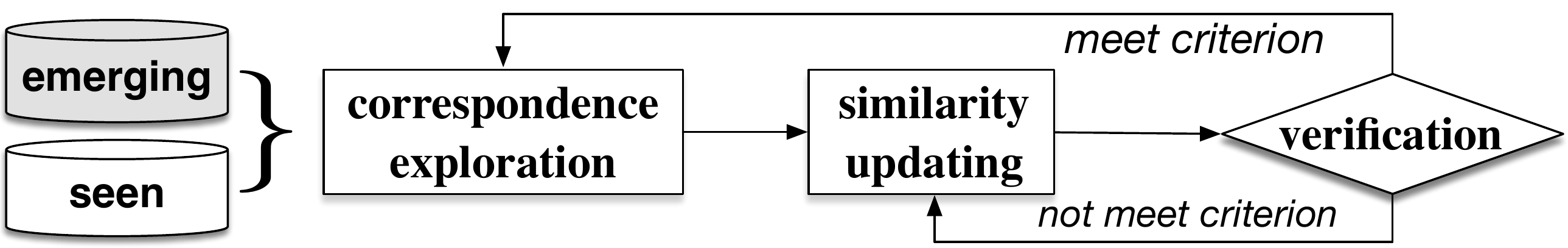}
\caption{Proposed self-learning paradigm to match the fully unsupervised data of an emerging modality using some already learned data in one seen modality.}
\label{fig:human-self-learning}
\end{figure}

\emph{Template-matching model}~\cite{anderson2005cognitive} is a major theory of pattern recognition in cognitive psychology, which explains the human visual recognition as a matching process between the incoming information and the seen templates stored in brain. Accordingly, we propose a self-learning paradigm for the cross-modality data matching from the perspective of cognitive psychology, as shown in Fig.~\ref{fig:human-self-learning}. First, the correspondence exploration is executed between data in seen modality (i.e., templates) and emerging modality (i.e., incoming information). Then, these subjectively conjectural correspondences are used as basic ``knowledge'' to update the similarity measurement at cognition level so as to give human an updated insight into the cross-modality data correspondence, followed by a verification procedure to determine whether the update meets a predetermined criterion. If not, the similarity updating continues; otherwise, it is leveraged to explore more potential data correspondences in the next learning iteration. This recurrent and exploratory self-learning paradigm can assist human in progressively ingesting and digesting the incremental novel visual information. Technically, we design \emph{self-reinforcing whole image matching} (SWIM) algorithm to simulate the aforementioned self-learning process. Within each SWIM iteration, first, we capitalize on the statistics of DPW distances of all cross-modality data pairs to explore potential data correspondence. Then, we devise a learnable LoFA to enable the local representation transfer between cross-modality data, guaranteeing the similarity evaluation in a compatible feature space, and develop a \emph{self-reinforcing local feature matching} (SLoMa) algorithm to iteratively optimize the LoFA. With the learnable weights of the LoFA, as the verification criterion, not fluctuating seriously, 
the correspondence exploration in the next SWIM iteration 
for more potentially matched data can be fulfilled. 
Obviously, SLoMa is implemented inside each SWIM iteration. Both the outer SWIM and the inner SLoMa learn from the self-reinforcing mechanism in behavioristics, and the data in the emerging modality are fully unlabeled; thus, we entitle the overall framework as SUM framework. 

\subsection{Feature Matrices}\label{fe} 
SUM is tailored for the feature representation maintaining 2D spatial information of raw images. Thus, we need to devise a feature encoder to embed images into \emph{feature matrices}, which acts as a high-level abstraction for low-level visual images. Convolutional neural network (CNN)~\cite{simonyan2014very} is a straightforward choice. Specifically, we use translation, shearing and scaling transformation to augment the data of the seen modality where only one template is available. The augmented data are used to train a CNN to classify these data in the seen modality into $N$ object classes. Once trained well, the CNN blocks prior to fully connected layers are frozen as a feature encoder. The modality-specific encoder embeds $\{\mathbf{x}^s_i\}_{i=1}^N$ and $\{\mathbf{x}^e_i\}_{i=1}^N$ into $\{\mathbf{S}_i\}_{i=1}^N$ and $\{\mathbf{E}_i\}_{i=1}^N$, respectively, by its convolutional, ReLU~\cite{nair2010rectified} and maxpooling layers, where $\mathbf{S}_i$, $\mathbf{E}_i \in \mathbb{R}^{H\times W \times C}$ are  feature tensors (or maps). The position-dependent vector along $C$ channels in a feature tensor is defined as the \emph{feature element}, and the feature tensors can be viewed as feature matrices in the form of $\mathbf{S}_i = [\mathbf{s}_{ihw}]^{H\times W}$ and $\mathbf{E}_i = [\mathbf{e}_{ihw}]^{H\times W}$, where feature elements $\mathbf{s}_{ihw},\mathbf{e}_{ihw}\in \mathbb{R}^{C}$ encode the local representation corresponding to their spatial receptive field in raw image.

\subsection{Dynamic Position Warping}\label{DPW}
DPW is designed to formulate the alignment between two 2D feature matrices and identify their structual correspondence. For notation brevity and conceptual generality, we omit the aforementioned subscript $i$ of $\mathbf{E}_i$ and $\mathbf{S}_i$, and use $\mathbf{S}=[\mathbf{s}_{hw}]^{H^s\times W^s}$ and $\mathbf{E}=[\mathbf{e}_{hw}]^{H^e\times W^e}$ to represent the two feature matrices to be aligned. We have not declared that $H^s$$=$$H^e$ and $W^s$$=$$W^e$ to show the scalability of the proposed DPW when the size of $\mathbf{E}$ and $\mathbf{S}$ are different, although their size as well as the image size in seen and emerging modality used in our experiments are identical. In this section, we detail the definition of the  proposed DPW algorithm and the involved mathematical calculation. Moreover, a toy matching example by DPW is also given to enhance comprehensibility.

\subsubsection{Definition}
Let $\mathbf{S}$ and $\mathbf{E}$ to be two feature matrices to be aligned, where feature elements $\mathbf{s}_{hw}$ and $\mathbf{e}_{hw}$ can be scalars or vectors with same dimension. The value space of $\mathbf{s}_{hw}$ and $\mathbf{e}_{hw}$ is denoted by $\mathcal{F}$. We give the following definition to formalize the notion of alignment between $\mathbf{S}$ and $\mathbf{E}$.

\textbf{Definition 1.} An $(H^s,W^s)$--$(H^e,W^e)$ hierarchical warping path (HiPa) $\mathbf{p}$ between two feature matrices $\mathbf{S}=[\mathbf{s}_{hw}]^{H^s\times W^s}$ and $\mathbf{E}=[\mathbf{e}_{hw}]^{H^e\times W^e}$ is a tree-based two-tuple path node set, $\mathbf{s}_{hw},\mathbf{e}_{hw}\in \mathbb{R}^{C}$, which is denoted as follows:
\begin{center}
\small
\begin{tikzpicture}[scale=0.96]
\centering
\node {$(H^s,W^s)$--$(H^e,W^e)$ HiPa: $\mathbf{p}$}
child {node {$p^{(1)}_1=(h^s_1,h_1^e)$}
child {node {$\ldots$}}
child [missing] {} 
}
child {node {$\ldots$}}
child {node {$p^{(1)}_l=(h^s_l,h^e_l)$}
child {node {$p^{(2)}_1=(w^s_1,w^e_1)$}}
child {node {$\ldots$}}
child {node {$p^{(2)}_{c_l}=(w^s_{c_l},w^e_{c_l})$}}
}
child {node {$\ldots$}}
child {node {$p^{(1)}_L=(h^s_L,h^e_L)$}
child [missing] {}
child {node {$\ldots$}}
};
\end{tikzpicture}
\normalsize
\end{center}
where $p^{(1)}_l$$=\ $$(h^s_l,h^e_l)\in [1$$:$$\ H^s]$$\times$$[1$$:$$\ H^e],\ p^{(2)}_k$$=\ $$(w^s_k, w^e_k) \in [1$$:$$\ W^s]$$\times$$[1$$:$$\ W^e]$ are two-tuple path nodes, $l\in [1$$:$$\ L]$, $k \in [1$$:$$\ c_l]$. In other words, $\mathbf{p}$ consists of a series of two-tuple path nodes and it defines a dense element-level alignment path between $\mathbf{S}$ and $\mathbf{E}$. Specifically, the first-level node $p^{(1)}_l=(h^s_l,h^e_l)$ indicates that the $h^s_l$-th row feature sequence of $\mathbf{S}$ (i.e., $\mathbf{S}(h^s_l)=(\mathbf{s}_{h^s_l1},\ldots,\mathbf{s}_{h^s_l W^s}) \in \mathbb{R}^{C \times W_s}$) is determined to match the $h^e_l$-th row feature sequence of $\mathbf{E}$ (i.e., $\mathbf{E}(h^e_l)=(\mathbf{e}_{h^e_l1},\ldots,\mathbf{e}_{h^e_l W^e}) \in \mathbb{R}^{C\times W_e}$), and the newly defined symbolic notation $p^{(2)}_k|\ p^{(1)}_l$$=$$(w^s_{k},w^e_{k})|(h^s_l,h^e_l)$ indicates that the feature element of $\mathbf{S}$ at coordinate ($h^s_l,w^s_{k}$) (i.e., $\mathbf{s}_{h^s_l w^s_{k}}$$\in$$\mathbb{R}^{C}$) is determined to match that of $\mathbf{E}$ at coordinate ($h^e_l,w^e_{k}$) (i.e., $\mathbf{e}_{h^e_l w^e_{k}}$$\in$$\mathbb{R}^{C}$). Note that the symbolic notation $\cdot|\ p^{(1)}_l$ represents the event $\cdot$ conditioned on first-level alignment $p^{(1)}_l$. In addition, $L$ represents the number of first-level path nodes in $\mathbf{p}$, and $c_l$ denotes the number of second-level path nodes conditioned on the specific first-level alignment described by $p^{(1)}_l$, where $l$ $\in$$[1:L]$. Actually, there are $\sum_{l=1}^{L} c_l$ pairs of aligned feature elements between $\mathbf{S}$ and $\mathbf{E}$ in total.
Moreover, as a HiPa, $\mathbf{p}$ must satisfy the following three conditions:

\noindent (a) Hierarchical boundary condition (HBC): 
\begin{equation}
  \left\{
   \begin{aligned}
   & p^{(1)}_1=(1,1),\ p^{(1)}_L=(H^s,H^e),\\
   & p^{(2)}_1|\ p^{(1)}_l =(1,1)|(h^s_l,h^e_l),\ \forall \ l \in [1, L],\\
   & p^{(2)}_{c_l}|\ p^{(1)}_l =(W^s,W^e)|(h^s_l,h^e_l),\ \forall \ l \in [1, L].
   \end{aligned}
   \right.
   \label{eq:HBC}
  \end{equation}

 \noindent (b) Hierarchical monotonicity condition (HMC):
 \begin{equation}
  \left\{
   \begin{aligned}
   & h^s_1\leq h^s_2\leq \ldots \leq h^s_L,\ h^e_1\leq h^e_2\leq \ldots \leq h^e_L\\
   & w^s_1\leq w^s_2\leq \ldots \leq w^s_{c_l}|\ p^{(1)}_l,\ \forall\ l \in [1,L],\\
   & w^e_1\leq w^e_2\leq \ldots \leq w^e_{c_l}|\ p^{(1)}_l,\ \forall \ l \in [1, L].\qquad  
   \end{aligned}
   \right.
   \label{eq:HMC}
  \end{equation}

\noindent (c) Hierarchical step size condition (HSC):
 \begin{equation}
  \left\{
   \begin{aligned}
   & p^{(1)}_{l+1}-p^{(1)}_l\in\{(1,0),(0,1),(1,1)\}, \forall\ l \in [1,L-1],\\
   & p^{(2)}_{k+1}-p^{(2)}_k|\ p^{(1)}_l\in \{(1,0)|\ p^{(1)}_l,(0,1)|\ p^{(1)}_l,(1,1)|\ p^{(1)}_l\},\\
   & \qquad\qquad\qquad\qquad\qquad\qquad\ \ \  \forall \ l \in [1, L], k\in [1,c_l-1].  
   \end{aligned}
   \right.
   \label{eq:HMC}
  \end{equation}

\begin{algorithm}[t!]
   \caption{Calculating the optimal HiPa}
   \label{algo:oHWP}
\begin{algorithmic}[1]
   \State {\bfseries Input:} Hierarchical accumulated distance matrix $\mathcal{D}\in \mathbb{R}^{H^s\times H^e}$ between $\mathbf{S}\ $$=$$[\mathbf{s}_{hw}]^{H^s\times W^s}$ and $\mathbf{E}\ $$=$$[\mathbf{e}_{hw}]^{H^e\times W^e}$.
   \State {\bfseries Output:} Optimal HiPa between the two feature matrices: $\mathbf{p}^*=\{p^{(1)*}_l, p^{(2)*}_k|\ p^{(1)*}_l, k=1,\ldots,c_l, l=1,\ldots,L\}$.

  \vskip 0.05in
   \noindent ----------- \emph{{\bfseries STEP 1:} Calculate the first-level path nodes} -----------
   \State  Let $p^{(1)*}_L=(h^s_L, h^e_L)=(H^s, H^e)$ and $l=L$.
   \Repeat
   \begin{equation}
   \begin{aligned}
   &\ \ \ \ p^{(1)*}_{l-1} = (h^s_{l-1}, h^e_{l-1})=\\
  &\ \left\{{}
   \begin{aligned}
    &(1, h^e_{l}-1),\ \mathrm{if}\ h^s_{l}=1,\\
    &(h^s_{l}-1, 1),\ \mathrm{if}\ h^e_{l}=1,\\
    &\mathop{\arg\min}\nolimits_{(i,j)\in F}\mathcal{D}(i,j), \mathrm{where}\ F=\{(h^s_{l}-1,h^e_l-1),\\
    &\qquad\qquad\qquad\ (h^s_{l}-1,h^e_l),(h^s_{l},h^e_l-1)\}, \ \mathrm{otherwise}.
   \end{aligned}
   \right.
  \end{aligned}
  \label{eq:first-level nodes}
  \end{equation}
   \State Let $l=l-1.$
   \vskip 0.02in
  \Until{$p^{(1)*}_l=(h_l^s,h_l^e)=(1,1)$}
  \vskip 0.02in
  \State Let $l=1$ and sequentially assign the incremental index numbers to $l+1,l+2,\ldots,L$, and then obtain first-level path nodes $p^{(1)*}_1,\ldots,p^{(1)*}_L $.

  \vskip 0.05in
   \noindent --------- \emph{{\bfseries STEP 2:} Calculate the second-level path nodes} ---------
  \For{$l=1$ to $L$}
    \State Let $ p^{(2)*}_{c_l}=(w^s_{c_l}, w^e_{c_l})=(W^s, W^e)$ and $k=c_l$.
   \Repeat
  \begin{equation}
  \begin{aligned}
  &\ \ \ \ \ \ \ \ \ \ p^{(2)*}_{k-1} = (w^s_{k-1},w^e_{k-1})= \\
  &\left\{
   \begin{aligned}
    &(1, w^e_{k}-1),\ \mathrm{if}\ w^s_{k}=1,\\
    &(w^s_{k}-1, 1),\ \mathrm{if}\ w^e_{k}=1,\\
    &\mathop{\arg\min}\nolimits_{(i,j)\in F}\mathcal{DTW}(\mathbf{S}(h^s_l,1:i),\mathbf{E}(h^e_l,1:j)), \mathrm{where}\\ 
    &\ \ \  F=\{(w^s_{k}-1,w^e_k-1),(w^s_{k}-1,w^e_l),(w^s_{k},w^e_k-1)\}.
   \end{aligned}
   \right.
   \end{aligned}
  \label{eq:second-level nodes}
  \end{equation}

   \noindent \small (\emph{Note that $\mathbf{S}(m,1:n)$$=$$(\mathbf{s}_{m1},\ldots,\mathbf{s}_{mn})$ denotes a sub-sequence of the $m$-th row feature sequence of $\mathbf{S}$, and analogous for $\mathbf{E}$. All values of $\mathcal{DTW}(\cdot,\cdot)$ above have been storaged during calculating the DPW distance as shown in Eq.~\ref{eq:DPW_distance_solve}.})
    \normalsize
      \vskip 0.02in
  \State Let $k=k-1$.
      \vskip 0.02in
    \Until{$p^{(2)*}_k=(w_k^s,w_k^e)=(1,1)$}
    \vskip 0.01in
     \State Let $k=1$ and sequentially assign the incremental index numbers to $k+1,k+2,\ldots,c_l$, and the obtain second-level path nodes $p^{(2)*}_1|\ p^{(1)*}_l,\ldots,p^{(2)*}_{c_l}|\ p^{(1)*}_l$.
    \EndFor

\end{algorithmic}
\end{algorithm}

Intuitively, HBC enforces that the first feature elements of $\mathbf{S}$ and $\mathbf{E}$ as well as the last feature elements of $\mathbf{S}$ and $\mathbf{E}$ are aligned to each other, it guarantee a entire coverage for alignment. HMC ensures the 2D structure-preserving property and HSC expresses a kind of continuity that no feature element in $\mathbf{S}$ and $\mathbf{E}$ can be omitted and no repeated element alignment can occur. In this way, we define the matching distance between $\mathbf{S}$ and $\mathbf{E}$ along $(H^s,W^s)$--$(H^e,W^e)$ HiPa $\mathbf{p}$ as $\mathbf{D}_\mathbf{p}(\mathbf{S},\mathbf{E})$:
\begin{equation}
\begin{aligned}
\mathbf{D}_\mathbf{p}(\mathbf{S},\mathbf{E})&=\mathbf{D}_{\{p^{(1)}_1,\ldots,p^{(1)}_L\}}(\mathbf{S},\mathbf{E})\\
&=\sum_{l=1}^{L}\mathbf{D}_{\{p^{(2)}_1,\ldots,p^{(2)}_{c_l}\}|\ p^{(1)}_l}(\mathbf{S}(h^s_l), \mathbf{E}(h^e_l))
\\
&=\sum\limits_{l=1}^{L}\sum\limits_{k=1}^{c_l}d(\mathbf{s}_{h^s_l w^s_k},\mathbf{e}_{h^e_l w^e_k}),
\end{aligned}
\label{eq:total_cost}
\end{equation}
where $d: \mathcal{F}\times \mathcal{F} \rightarrow \mathbb{R}_{\ge 0}$ denotes the distance measurement between feature elements (we adopt Euclidean distance). 

\textbf{Definition 2.} Let $\mathrm{P}(\mathbf{S},\mathbf{E})$ to be the HiPa collection which includes all feasible HiPas between the two feature matrices, $\mathbf{S}$ and $\mathbf{E}$, and let $\mathbf{p}^*$ to be a HiPa between $\mathbf{S}$ and $\mathbf{E}$ from this HiPa collection, $\mathbf{p}^*\in \mathrm{P}(\mathbf{S},\mathbf{E})$, which satisfies
\begin{equation}
\mathbf{D}_{\mathbf{p}^*}(\mathbf{S},\mathbf{E})=\mathop{\min}\{\mathbf{D}_{\mathbf{p}}(\mathbf{S},\mathbf{E})\ |\ \mathbf{p} \in \mathrm{P}(\mathbf{S},\mathbf{E})\}.
\end{equation}
Then, $\mathbf{D}_{\mathbf{p}^*}(\mathbf{S},\mathbf{E})$ is entitled as {\it DPW distance} between $\mathbf{S}$ and $\mathbf{E}$, i.e., $\mathcal{DPW}(\mathbf{S},\mathbf{E}) = \mathbf{D}_{\mathbf{p}^*}(\mathbf{S},\mathbf{E})$, and the corresponding $\mathbf{p}^{*}$ is entitled as {\it optimal HiPa} between $\mathbf{S}$ and $\mathbf{E}$.

\subsubsection{Mathematical Calculation}

In this part, we discuss how to compute the DPW distance as well as the corresponding optimal HiPa between two feature matrices, $\mathbf{S}=[\mathbf{s}_{hw}]^{H^s\times W^s}$ and $\mathbf{E}=[\mathbf{e}_{hw}]^{H^e\times W^e}$. For DPW distance calculation, we develop an $O(H^sW^sH^eW^e)$ algorithm via dynamic programming:
\begin{equation}
  \left\{
   \begin{aligned}
   &\mathcal{DPW}(\mathbf{S},\mathbf{E})=\mathcal{D}(H^s,H^e),\\
   &\mathcal{D}(h^s,1)=\sum\nolimits_{i=1}^{h^s}\mathcal{DTW}(\mathbf{S}(i),\mathbf{E}(1)),\ h^s\in[1:H^s],\\
   &\mathcal{D}(1,h^e)=\sum\nolimits_{i=1}^{h^e}\mathcal{DTW}(\mathbf{S}(1),\mathbf{E}(i)),\ h^e \in[1:H^e],
   \\
   &\mathcal{D}(h^s,h^e) = \mathop{\min}\{\mathcal{D}(h^s-1,h^e-1),\mathcal{D}(h^s-1,h^e), \\
   &\qquad \qquad \quad \ \ \mathcal{D}(h^s,h^e-1)\}+\mathcal{DTW}(\mathbf{S}(h^s),\mathbf{E}(h^e)),\\
  &\qquad\qquad\quad\ \  
      h^s\in[2:H^s],\ h^e\in[2:H^e], \\
   \end{aligned}
   \right.
   \label{eq:DPW_distance_solve}
  \end{equation}
where $\mathcal{D}$ $ \in$ $ \mathbb{R}^{H^s\times H^e}$ is  hierarchical accumulated distance matrix which stores some intermediate computing results, and $\mathcal{DTW}(\cdot,\cdot)$ represents DTW distance which can be solved by the off-the-shelf DTW algorithm~\cite{sakoe1978dynamic} ({\color{blue}see our Appendix}). 

After obtaining the DPW distance by the forward computation above, we perform a backward-deduction to compute the corresponding optimal HiPa $\mathbf{p}^*$ between $\mathbf{S}$ and $\mathbf{E}$, whose detailed algorithm is provided in Alg.~\ref{algo:oHWP}. It is worth noting that the backward-deduction process entails some intermediate computing results including the hierarchical accumulated distance matrix $\mathcal{D}$ and the values of DTW distance between row feature sub-sequence of $\mathbf{S}$ and $\mathbf{E}$. These intermediate results can be storaged in the memory during computing the DPW distance to avoid the redundant computation. Furthemore, we provide a  theoretical proof for the optimality of $\mathbf{p}^{*}$ in our Appendix.

\begin{figure}[t!]
\centering
\subfloat[Gray image $\mathbf{S}$ of size $8\times 8$.]{
 \includegraphics[width=0.8\linewidth]{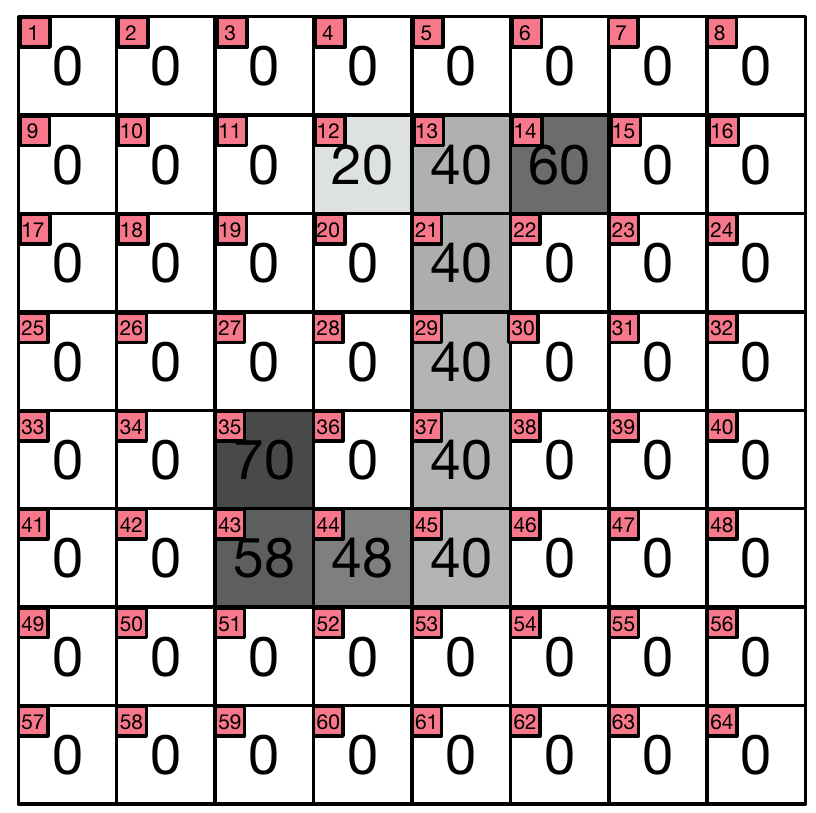}
}
\hspace{1.5em}
\subfloat[Gray image $\mathbf{E}$ of size $8\times 8$.]{
 \includegraphics[width=0.8\linewidth]{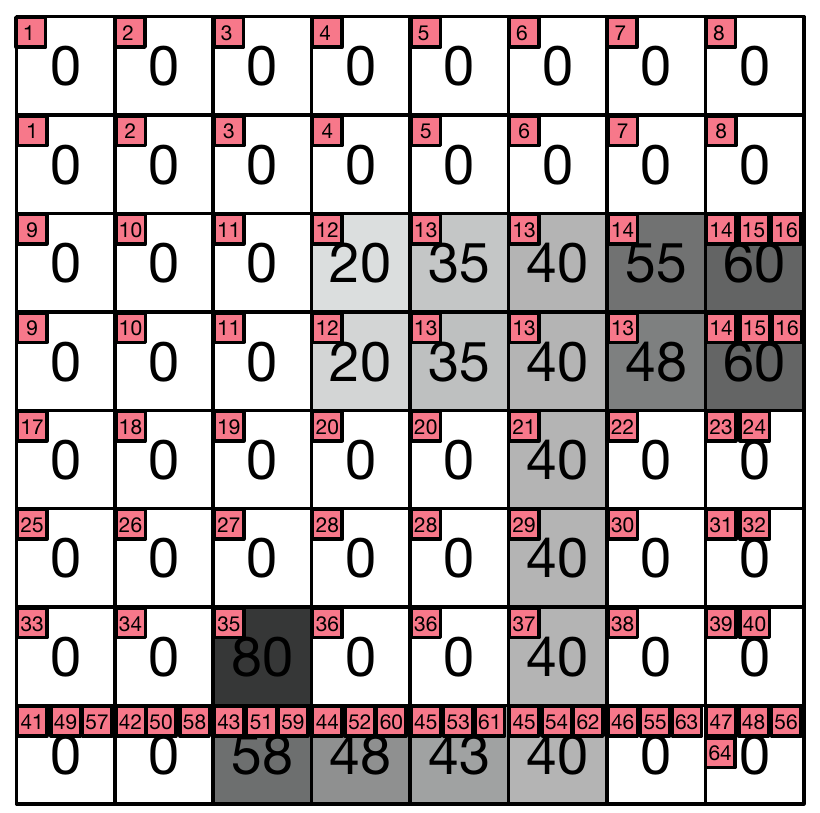}
}
\caption{A toy matching example for DPW. The gray values are marked inside the corresponding pixel points. The indices of $\mathbf{S}$'s pixel points are labeled with numbers in the red boxes. The assignments of $\mathbf{S}$'s pixel points in $\mathbf{E}$ are depicted with the same numbers in red boxes in (b). If using Euclidean Distance as the distance measurement for two pixel points, the DPW distance $\mathcal{DPW}(\mathbf{S},\mathbf{E})=654$. However, the naive point-wise distance $||\mathbf{S}-\mathbf{E}||_{1}=1118$.}
\label{fig:toy_example}
\end{figure}

\subsubsection{Toy Matching Example by DPW}
To better understand the proposed DPW, a toy matching example is reported in Fig.~\ref{fig:toy_example}. $\mathbf{S}$ and $\mathbf{E}$ represent two 2D gray images in which each pixel is associated with a real-value number.
Intuitively, the gray image $\mathbf{E}$ is a derivant by performing some transformations on gray image $\mathbf{S}$, such as translation, stretching or compression both vertically and horizontally. Actually, the two gray images  all implicitly contain the ``J''-like objects. As we can see, the proposed DPW is able to describe the correspondence between $\mathbf{S}$ and $\mathbf{E}$. In detail, the correspondences between $\mathbf{S}$ and $\mathbf{E}$ explored by DPW are illustrated by the index numbers in the red boxes. The two individual positions in $\mathbf{S}$ and $\mathbf{E}$ which are determined to match each other have been marked using the same index numbers. Moreover, the DPW distance can be regraded as the distance measuring the extent to which the two gray images match each other. If using the Euclidean distance as the distance measurement between two pixel elements, the final DPW distance between $\mathbf{S}$ and $\mathbf{E}$ is 654. However, the naive point-wise distance $||\mathbf{S}-\mathbf{E}||_{1}$ is 1118, which struggles to describe the high-level similarity between the two gray images. 

\begin{figure*}[t!]
\centering
\includegraphics[width=1\linewidth]{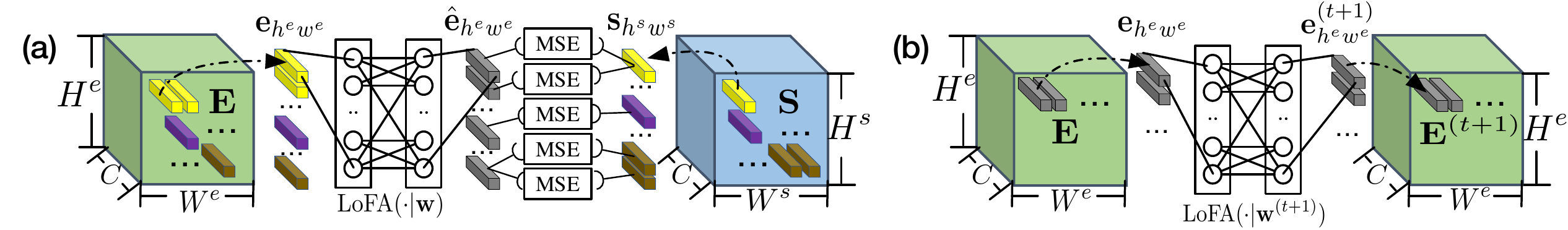}
\caption{Diagram for (a) ``optimize'' and (b) ``adapt'' in SLoMa. In (a), the strips of $\mathbf{E}$ and $\mathbf{S}$ in same color represent matched feature elements evaluated by DPW. In (b), the strips also represent feature elements and the adapted feature elements in $\mathbf{E}^{(t+1)}$ maintain their original positions in $\mathbf{E}$.}
\label{fig:LoFA}
\end{figure*}

\subsection{Local Feature Adapter}\label{subsec:LoFA}
An important prerequisite for using DPW to match two feature matrices is a compatible and comparable feature space that their feature elements populate. Therefore, we design the LoFA to counteract the local representation discrepancies between two cross-modality congener objects. Mimicking the human's similarity-association mechanism in psychology, the LoFA aims at achieving element-to-element conversion from the emerging modality to the seen modality, which explicitly embeds a feature element of emerging modality into an adapted modality similar to that of the seen modality. Technically, the Multi-Layer Perceptron~\cite{bishop1995neural} (MLP) is a concise choice for LoFA, though it can be substituted by other sophisticated embedding frameworks. The tailored LoFA holds two fundamental characteristics:
(1) Both its input and output are feature elements with  same dimension (i.e., $C$-dimensional).
(2) Its learnable parameters $\mathbf{w}$ are shared across all local feature elements because the parametric LoFA is expected to maintain good generalization ability and low model complexity.

\begin{algorithm}[tb!]
   \caption{SLoMa}
   \label{algo:SrLFM}
\begin{algorithmic}[1]
   \State {\bfseries Input:} Two sets of potentially matched feature matrices in the seen and emerging modality $\{\mathbf{S}_{k_i}\}_{i=1}^{n}$ and $\{\mathbf{E}_{l_i}\}_{i=1}^{n}$, where $k_i, l_i, n \in$$[1$\ :\ $N]$, $l_1$$\neq$$l_2$$\neq$$\cdots$$\neq$$l_n$ and $\mathbf{E}_{l_i}$ is determined to match $\mathbf{S}_{k_i}$ for $i\in[1:n]$; The iteration error threshold $\epsilon$.
   \State {\bfseries Initialize:} $\mathbf{w}=\mathbf{w}^{(0)}$ (random); $\mathbf{E}^{(0)}_{l_i} = \mathrm{LoFA}(\mathbf{E}_{l_i}|\mathbf{w}^{(0)})$, for $i\in [1:n]$; $t=0$.
   \Repeat
     \For{$i=1$ to $n$}

    \State Search the optimal HiPa $\mathbf{p}^{(t)*}_{i}$ between $\mathbf{S}_{k_i}$ and $\mathbf{E}^{(t)}_{l_i}$ by the DPW algorithm (see Eq.~\ref{eq:DPW_distance_solve} and Algorithm.~\ref{algo:oHWP}):

    \State $\mathbf{p}^{(t)*}_{i}$$=$$\mathop{\arg \min}_{\mathbf{p}\in \mathrm{P}(\mathbf{S}_{k_i},\mathbf{E}^{(t)}_{l_i})
    } \mathbf{D}_{\mathbf{p}}(\mathbf{S}_{k_i},\mathbf{E}^{(t)}_{l_i}).$\ $\triangleright${\color{blue}\emph{Match}}
    \EndFor

  \State Retrain the $\mathrm{LoFA}(\cdot|\mathbf{w})$ to shorten the matching distance between $\mathbf{S}_{k_i}$ and $\mathrm{LoFA}(\mathbf{E}_{l_i}|\mathbf{w})$ on HiPa $\mathbf{p}^{(t)*}_{i}$ to obtain the updated $\mathbf{w}^{(t+1)}$: \qquad\qquad\qquad\qquad\ $\triangleright${\color{blue}\emph{Optimize}}

  \State $\mathbf{w}^{(t+1)}$$=$$\mathop{\arg\min}_{\mathbf{w}} \frac{1}{n}\sum_{i=1}^{n}\mathbf{D}_{\mathbf{p}^{(t)*}_{i}}(\mathbf{S}_{k_i},\mathrm{LoFA}(\mathbf{E}_{l_i}|\mathbf{w}))$.

  \State $t++$; $\mathbf{E}^{(t)}_{l_i}$$=$$\mathrm{LoFA}(\mathbf{E}_{l_i}|\mathbf{w}^{(t)}), i$$\in$$[1:n]$. \qquad \quad $\triangleright${\color{blue}\emph{Adapt}}
    \Until{$||\mathbf{w}^{(t)}-\mathbf{w}^{(t-1)}||_{2}\leq \epsilon$}
    \State {\bfseries Output:} $\mathbf{w}^{*}=\mathbf{w}^{(t)}$, local feature adapter $\mathrm{LoFA}(\cdot|\mathbf{w}^*)$. 
\end{algorithmic}
\end{algorithm}

\begin{algorithm}[tb]
   \caption{SWIM}
   \label{algo:SrWIM}
\begin{algorithmic}[1]
   \State {\bfseries Input:} Two sets of feature matrices in seen and emerging modality to be matched, $\{\mathbf{S}_{i}\}_{i=1}^{N}$ and $\{\mathbf{E}_{i}\}_{i=1}^{N}$; The iteration error threshold $\epsilon$; The exploration step size $\alpha$.
   \vskip 0.02in
   \State {\bfseries Initialize:}  $T=0$; $n_0 =0$; $\{\mathbf{E}_{l_i}\}_{i=1}^{0}=\emptyset$;  $\mathrm{LoFA}(\mathbf{X}|\mathbf{w}^{*}_0) = \mathbf{X}$ is an identical function designed only for the unified description of this algorithm.
   \Repeat

   \State $T++$; $n_T = \alpha T$. \qquad\qquad\qquad\qquad\qquad\qquad\ \ $\triangleright${\color{blue}\emph{Update}}
   \For{$n=1$ to $n_T$}  \qquad\ $\triangleright${\color{blue}\emph{Build $\{\mathbf{S}_{k_i}\}_{i=1}^{n_T}$ and $\{\mathbf{E}_{l_i}\}_{i=1}^{n_T}$}}
      \vskip 0.02in 
   \State $\mathcal{E}=\{\mathbf{E}_i\}_{i=1}^{N}-\{\mathbf{E}_{l_i}\}_{i=1}^{n-1},$
   \vskip 0.02in
   \State $\mathbf{E}_{l_n}$$=$$\mathop{\arg \min}\limits_{\mathbf{E}\in\mathcal{E}}\mathop{\min}\limits_{\mathbf{S}\in\{\mathbf{S}_{i}\}_{i=1}^{N}} $$\mathcal{DPW}(\mathbf{S}, \mathrm{LoFA}(\mathbf{E}|\mathbf{w}_{T-1}^{*})),$

   \vskip 0.02in
   \State $\mathbf{S}_{k_n}$$=$$\mathop{\arg\min}\limits_{\mathbf{S}\in\{\mathbf{S}_{i}\}_{i=1}^{N}}\mathcal{DPW}(\mathbf{S}, \mathrm{LoFA}(\mathbf{E}_{l_n}|\mathbf{w}_{T-1}^{*}))\}.$
   \EndFor 

  \State $\mathrm{LoFA}(\cdot|\mathbf{w}^{*}_T)$$\leftarrow$$\mathrm{SLoMa}(\{\mathbf{S}_{k_i}\}_{i=1}^{n_T}, \{\mathbf{E}_{l_i}\}_{i=1}^{n_T}, \epsilon)$.\ $\triangleright${\color{blue}\emph{Alg.~\ref{algo:SrLFM}}}

    \Until{$n_T = N$}

    \State {\bfseries Output:} Two matched sets $\{\mathbf{S}_{k_i}\}_{i=1}^{N}$ and $\{\mathbf{E}_{l_i}\}_{i=1}^{N}$ where $\mathbf{E}_{l_i}$ is determined to match $\mathbf{S}_{k_i}$. 
\end{algorithmic}
\end{algorithm}

\subsection{Self-reinforcing Local Feature Matching}
In the strictest sense, $\mathrm{LoFA}(\cdot|\mathbf{w})$ should be optimized via explicit element-level correspondence between the matched cross-modality objects, whereas no ground-truth image or element correspondence is available due to the fully unsupervised modality. We exploit an iterative algorithm, SLoMa, to optimize LoFA in a self-reinforcing paradigm associated with DPW. Specifically, we denote by $\{\mathbf{S}_{k_i}\}_{i=1}^{n}$ and $\{\mathbf{E}_{l_i}\}_{i=1}^{n}$ the potentially matched subsets of the seen and emerging modalities, respectively, in the current SWIM iteration (SWIM is described in Sec.~\ref{sec:SWIM}), where $k_i, l_i$ $\in$$[1:\ N]$, $l_1\neq l_2\neq \cdots \neq l_n$, $\mathbf{E}_{l_i}$ is determined to match $\mathbf{S}_{k_i}$, and $n$ $\in\ $$ [1:N]$ is the number of potentially matched feature matrices in current SWIM iteration. SLoMa is summarized in Alg.~\ref{algo:SrLFM}.  


\subsubsection{Optimization}
For simplicity, we clarify in Fig.~\ref{fig:LoFA}\emph{A} the optimization step of Alg.~\ref{algo:SrLFM} using two potentially matched feature matrices $\mathbf{E}=[\mathbf{e}_{h^e w^e}]^{H^e\times W^e}$ and $\mathbf{S}=[\mathbf{s}_{h^sw^s}]^{H^s\times W^s}$, where $\mathbf{e}_{h^ew^e},\mathbf{s}_{h^sw^s}$$\in\ $$ \mathbb{R}^{C}$. In this case, the optimization step reasonably reduces to 
\begin{equation}
\mathbf{w}^{(t+1)}=\mathop{\arg\min}_{\mathbf{w}} \mathbf{D}_{\mathbf{p}^{(t)*}}(\mathbf{S},\mathrm{LoFA}(\mathbf{E}|\mathbf{w})),
\end{equation}
where $\mathbf{p}^{(t)*}$ is the optimal HiPa between $\mathbf{S}$ and $\mathbf{E}^{(t)}$$ = $$\mathrm{LoFA(\mathbf{E}|\mathbf{w}^{(t)})}$. Especially, $\mathbf{p}^{(t)*}$ defines a series of positional correspondences between $\mathbf{S}$ and $\mathbf{E}$, as depicted in Fig.~\ref{fig:LoFA}\emph{A} by colored strips.\footnote{Actually, $\mathbf{p}^{(t)*}$ defines a series of positional correspondences between $\mathbf{S}$ and $\mathbf{E}^{(t)}$, but the correspondences can be migrated to $\mathbf{S}$ and $\mathbf{E}$ since the LoFA that converts $\mathbf{E}$ into $\mathbf{E}^{(t)}$ is agnostic to the position.} Assuming element $\mathbf{e}_{h^ew^e}$ is matched with element $\mathbf{s}_{h^sw^s}$, the LoFA is expected to adapt $\mathbf{e}_{h^ew^e}$ into a more appropriate feature space wherein the adapted feature element $\hat{\mathbf{e}}_{h^ew^e}\in \mathbb{R}^C$ is close to $\mathbf{s}_{h^sw^s}$.  Considering the DPW distance is computed by the Eucilidean distance, mean squared error (MSE)~\cite{lehmann2006theory} is selected as the metric loss, and it acts upon on each pair of matched feature elements to fulfill the aforementioned optimization objective.

\subsubsection{Adaptation}
Once trained well, LoFA is fixed to perform local feature adaptation for $\{\mathbf{E}_{l_i}\}_{i=1}^n$. For a feature matrix $\mathbf{E}=[\mathbf{e}_{h^e w^e}]^{H^e\times W^e}\in \{\mathbf{E}_{l_i}\}_{i=1}^n$, as shown in Fig.~\ref{fig:LoFA}\emph{B},  the adaptation operates on each feature element $\mathbf{e}_{h^e w^e} \in \mathbb{R}^C$ to form the adapted feature matrix $\mathbf{E}^{(t+1)}=[\mathbf{e}^{(t+1)}_{h^ew^e}]^{H^e\times W^e}$, where $\mathbf{e}^{(t+1)}_{h^ew^e} \in \mathbb{R}^C$. Specifically, the adaptation operation only re-encodes the content of feature element and does not change its original position in $\mathbf{E}$. We believe the adapted feature matrix $\mathbf{E}^{(t+1)}$ by $\mathrm{LoFA}(\cdot|\mathbf{w}^{(t+1)})$ exists in a more compatible feature space with $\mathbf{S}$ since LoFA is enhanced gradually by diminishing the local representation gaps between the seen and emerging modality. Meanwhile, it supports the rationality of using DPW to evaluate the similarities between $\mathbf{E}$ and $\mathbf{S}$.

\subsubsection{Theoretical Reinterpretation for SLoMa}
The EM algorithm is an elaborate technique for finding the maximum likelihood estimate of the parameters of a distribution from a given data set when the data is incomplete or has missing values~\cite{moon1996expectation,bilmes1998gentle}. EM algorithm is amenable for the problems wherein some missing values exist among data caused by the fault of data acquisition or the limitation of observation process, as well as the problems wherein the optimization of objective function is analytically difficult but it can be simplified by introducing some additional latent variables. Specifically, let $\theta^{t}$ be the estimated parameter at the $t$-iteration, $\mathbf{x}$ be the visible variable and $\mathbf{y}$ be the latent variable. EM algorithm iterates between two steps. the E-step computes the $Q$-function:
\begin{equation}
Q(\theta_{t},\theta)= \sum_{\mathbf{y}}p(\mathbf{y}|\mathbf{x},\theta_{t})\log p(\mathbf{x},\mathbf{y}|\theta)=E[\log p(\mathbf{x},\mathbf{y}|\theta)|\mathbf{x},\theta_{t}],
\end{equation}
and the M-step maximizes the $Q$-function with respect to $\theta$ so as to get the updated $\theta_{t+1}$:
\begin{equation}
\theta_{t+1} \leftarrow \mathop{\arg\max}_{\theta}Q(\theta_{t},\theta).
\end{equation}
In many real-world applications, however, the optimal $\theta_{t+1}$ in above M-step is hard to search. Thus, a more general operation in M-step is to find a suitable $\theta_{t+1}$ such that
\begin{equation}
Q(\theta_{t},\theta_{t+1})>Q(\theta_{t},\theta_{t}),
\label{eq:10}
\end{equation}
which also ensures the convergence of the EM algorithm.

The proposed LoFA is designed to be optimized by SLoMa, which alternates between match step and optimization step as described in Algorithm~\ref{algo:SrLFM}. Next, an EM-based theoretical analysis is given for better comprehension of the SLoMa algorithm. 

For simplicity, here we focus on only one pair of potentially matched cross-modality feature matrices $\mathbf{S}_{k_i}$ and $\mathbf{E}_{l_i}$. Furthermore, the two feature matrices $\mathbf{S}_{k_i}$ and $\mathbf{E}_{l_i}$ are regareded as visible variables, and the weights of LoFA $\mathbf{w}$  are parameters to be estimated. In particular, the HiPa $\mathbf{p}_i$ between $\mathbf{S}_{k_i}$ and $\mathbf{E}_{l_i}$ is set as the latent variable. Because the $\mathrm{LoFA}$ only re-encodes the context of each feature element of  and do not change the position of feature element. Hence, the size of $\mathrm{LoFA}(\mathbf{E}_{l_i}|\mathbf{w})$ always keep the same with that of $\mathbf{E}_{l_i}$. Thus, the HiPa collection between $\mathbf{S}_{k_i}$ and $\mathbf{E}_{l_i}$ is equivalent to that between $\mathbf{S}_{k_i}$ and the adapted emerging feature matrix $\mathrm{LoFA}(\mathbf{E}_{l_i}|\mathbf{w})$, which is fomulated as:
\begin{equation}
\mathrm{P}(\mathbf{S}_{k_i},\mathbf{E}_{l_i})=\mathrm{P}(\mathbf{S}_{k_i},\mathrm{LoFA}(\mathbf{E}_{l_i}|\mathbf{w})).
\end{equation}
We make an assumption that the probability distribution of $\mathbf{p}_i$ obeys the following softmax-like form:
\begin{equation}
Pr(\mathbf{p}_i| \mathbf{S}_{k_i},\mathbf{E}_{l_i})=\frac{\exp(-\alpha \mathbf{D}_{\mathbf{p}_i}(\mathbf{S}_{k_i},\mathbf{E}_{l_i}))}{\sum_{\hat{\mathbf{p}_i}\in \mathrm{P}(\mathbf{S}_{k_i},\mathbf{E}_{l_i})}\exp(-\alpha \mathbf{D}_{\hat{\mathbf{p}}_i}(\mathbf{S}_{k_i},\mathbf{E}_{l_i}))},
\label{eq:12}
\end{equation}
where $\alpha$ is positive scale constant. This probability distribution implies a negative correlation between the matching distance along one HiPa and the probability that this HiPa is chosen. Specially, the maximum probability occurs in the optimal HiPa $\mathbf{p}^*_i$ since $\mathbf{D}_{\mathbf{p}^*_i}(\mathbf{S}_{k_i},\mathbf{E}_{l_i})$ is the minimum matching distance, \emph{i.e.,} DPW distance. 
Therefore, given the previously estimated parameter $\mathbf{w}^{(t)}$, the $Q$ function is formed as:
\begin{equation}
\begin{aligned}
&Q(\mathbf{w}^{(t)},\mathbf{w})\\
=&\sum_{\mathbf{p}_i\in \mathrm{P}(\mathbf{S}_{k_i},\mathbf{E}_{l_i})}Pr(\mathbf{p}_i|\mathbf{S}_{k_i},\mathbf{E}_{l_i},\mathbf{w}^{(t)})\log Pr(\mathbf{S}_{k_i},\mathbf{E}_{l_i},\mathbf{p}_i|\mathbf{w})\\
=&\sum_{\mathbf{p}_i\in \mathrm{P}(\mathbf{S}_{k_i},\mathbf{E}_{l_i}^{(t)})}Pr(\mathbf{p}_i|\mathbf{S}_{k_i},\mathbf{E}_{l_i}^{(t)})\log Pr(\mathbf{S}_{k_i},\mathbf{E}_{l_i},\mathbf{p}_i|\mathbf{w}),\\
\end{aligned}
\end{equation}
where $\mathbf{E}_{l_i}^{(t)}=\mathrm{LoFA}(\mathbf{E}_{l_i}|\mathbf{w}^{(t)})$. In addition, by the conditional probability formula we can get
\begin{equation}
Pr (\mathbf{S}_{k_i}, \mathbf{E}_{l_i},\mathbf{p}_i|\mathbf{w}) = Pr(\mathbf{p}_i|\mathbf{S}_{k_i}, \mathbf{E}_{l_i},\mathbf{w})Pr(\mathbf{S}_{k_i}, \mathbf{E}_{l_i}|\mathbf{w}).
\label{eq:14}
\end{equation}
Since the visible variables $\mathbf{S}_{k_i}$ and $\mathbf{E}_{l_i}$ are irrelevant to the parameter $\mathbf{w}$, the $Pr(\mathbf{S}_{k_i}, \mathbf{E}_{l_i}|\mathbf{w})$ is actually a constant to $\mathbf{w}$. Consequently, the $Q$-function in Eq.~\ref{eq:14} reduces into the following form:
\begin{equation}
\begin{aligned}
&Q(\mathbf{w}^{(t)},\mathbf{w}) \\
= &\sum_{\mathbf{p}_i\in P(\mathbf{S}_{k_i},\mathbf{E}_{l_i}^{(t)})}Pr(\mathbf{p}_i|\mathbf{S}_{k_i},\mathbf{E}^{(t)}_{l_i})\log Pr(\mathbf{p}_i|\mathbf{S}_{k_i},\mathbf{E}_{l_i},\mathbf{w}).
\end{aligned}
\label{eq:15}
\end{equation}

\begin{figure*}[t]
\centering
\includegraphics[width=1.0\linewidth]{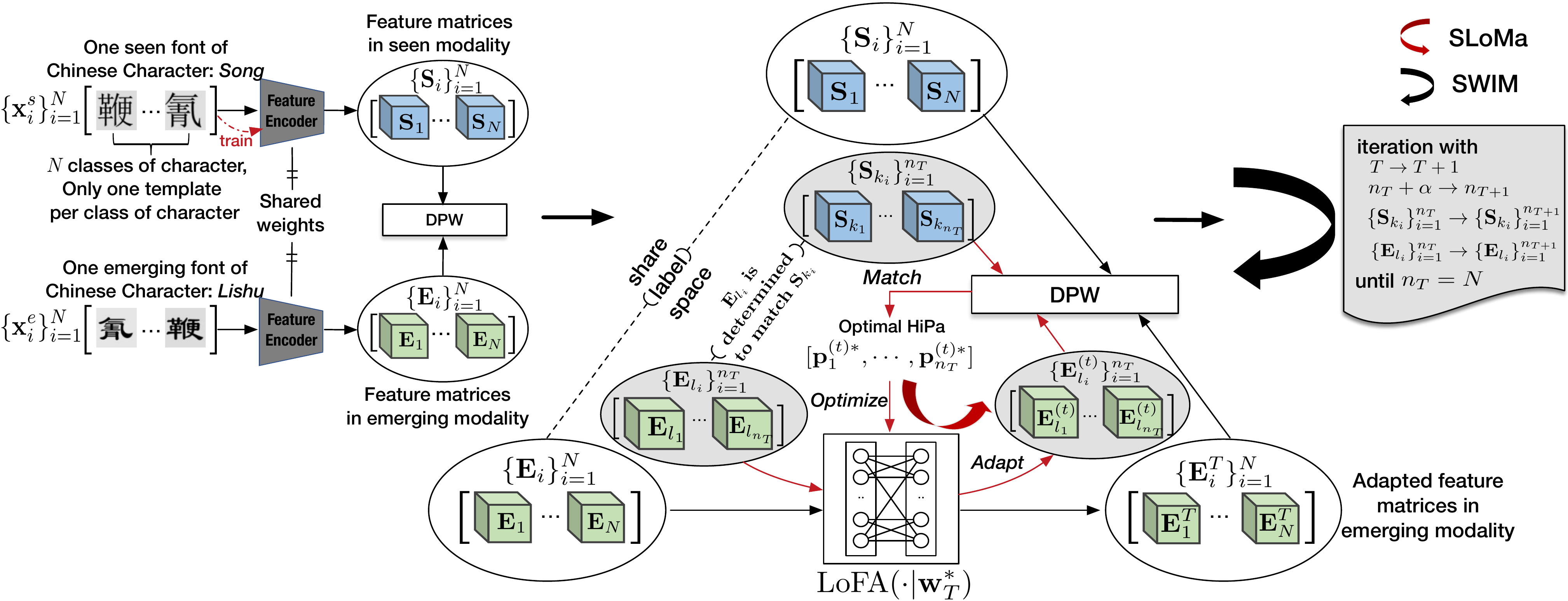}
\caption{Overall framework of SUM. SLoMa (\emph{red whirl arrow}) operates inside each SWIM iteration (\emph{black whirl arrow}). The subsets $\{\mathbf{S}_{k_i}\}_{i=1}^{n_T}$ and $\{\mathbf{E}_{l_i}\}_{i=1}^{n_T}$ in SWIM iteration $T$ are formed by absorbing the potentially matched cross-modality feature matrices in $\{\mathbf{S}_{i}\}_{i=1}^{N}$ and $\{\mathbf{E}_{i}\}_{i=1}^{N}$ according to the statistics of the DPW distances between $\{\mathbf{S}_{i}\}_{i=1}^{N}$ and the set of adapted feature matrices in the emerging modality $\{\mathbf{E}_i^T\}_{i=1}^N$, where $\mathbf{E}_i^T=\mathrm{LoFA}(\mathbf{E}_i|\mathbf{w}_T^{*}),\ i\in[1:N]$ (see Alg.~\ref{algo:SrWIM}).}
\label{fig:pipeline}
\end{figure*}

As mentioned above, the next M-step involves the maximization of the above $Q$-function with respect to $\mathbf{w}$. However, it is unrealistic to enumerate all feasible HiPas in $\mathrm{P}(\mathbf{S}_{k_i},\mathbf{E}_{l_i}^{(t)})$  and evaluate the probability for each HiPa, because the HiPa alternatives are numerous and its total number increases in exponential order with the size of feature matrix growing. Therefore, we switch to the following Hard-EM style to approximate the Eq.~\ref{eq:15}:
\begin{equation}
Q(\mathbf{w}^{(t)},\mathbf{w})\approx Pr(\mathbf{p}^{(t)*}_i|\mathbf{S}_{k_i},\mathbf{E}_{l_i}^{(t)})\log Pr(\mathbf{p}^{(t)*}_i|\mathbf{S}_{k_i},\mathbf{E}_{l_i},\mathbf{w}),
\label{eq:16}
\end{equation}
where
\begin{equation}
\mathbf{p}^{(t)*}_i = \mathop{\arg\max}_{\mathbf{p}_i\in \mathrm{P}(\mathbf{S}_{k_i},\mathbf{E}_{l_i}^{(t)})} Pr(\mathbf{p}_i|\mathbf{S}_{k_i},\mathbf{E}_{l_i}^{(t)}).
\end{equation}
In fact, the $\mathbf{p}^{(t)*}_i$ is the optimal HiPa between $\mathbf{S}_{k_i}$ and $\mathbf{E}_{l_i}^{(t)}$, which can be calculated by the DPW algorithm and the Alogorithm~\ref{algo:oHWP}. According to Eq.~\ref{eq:12}, we have:
\begin{equation}
\begin{aligned}
&Pr(\mathbf{p}^{(t)*}_i|\mathbf{S}_{k_i},\mathbf{E}_{l_i},\mathbf{w})\\
=&Pr(\mathbf{p}^{(t)*}_i|\mathbf{S}_{k_i},\mathrm{LoFA}(\mathbf{E}_{l_i}|\mathbf{w}))\\
=&\frac{\exp(-\alpha \mathbf{D}_{\mathbf{p}^{(t)*}_i}(\mathbf{S}_{k_i},\mathrm{LoFA}(\mathbf{E}_{l_i}|\mathbf{w}))}{\sum_{\hat{\mathbf{p}}_i\in \mathbf{P}(\mathbf{S}_{k_i},\mathbf{E}_{l_i})}\exp(-\alpha \mathbf{D}_{\hat{\mathbf{p}}_i}(\mathbf{S}_{k_i},\mathrm{LoFA}(\mathbf{E}_{l_i}|\mathbf{w}))}.
\end{aligned}
\end{equation}
Especially, the term of $Pr(\mathbf{p}^{(t)*}_i|\mathbf{S}_{k_i},\mathbf{E}_{l_i}^{(t)})$ in Eq.~\ref{eq:16} is irrelevant with respect to $\mathbf{w}$. Thus, the M-step involving the maximization of the $Q$-function in Eq.~\ref{eq:16} is simplified to:
\begin{equation}
\begin{aligned}
&\mathop{\max}_\mathbf{w} Q(\mathbf{w}^{(t)},\mathbf{w})
\sim \mathop{\max}_{\mathbf{w}} \log Pr(\mathbf{p}^{(t)*}_i|\mathbf{S}_{k_i},\mathbf{E}_{l_i},\mathbf{w})\\
=&\mathop{\max}_\mathbf{w}\log \frac{\exp(-\alpha \mathbf{D}_{\mathbf{p}^{(t)*}_i}(\mathbf{S}_{k_i},\mathrm{LoFA}(\mathbf{E}_{l_i}|\mathbf{w}))}{\sum_{\hat{\mathbf{p}}_i\in \mathbf{P}(\mathbf{S}_{k_i},\mathbf{E}_{l_i})}\exp(-\alpha \mathbf{D}_{\hat{\mathbf{p}}_i}(\mathbf{S}_{k_i},\mathrm{LoFA}(\mathbf{E}_{l_i}|\mathbf{w}))}\\
\sim & \mathop{\min}_\mathbf{w} \Big\{\underbrace{\alpha \mathbf{D}_{\mathbf{p}^{(t)*}_i}(\mathbf{S}_{k_i},\mathrm{LoFA}(\mathbf{E}_{l_i}|\mathbf{w})}_{\mathcal{L}_1(\mathbf{w})}\\
+&\underbrace{\log \sum_{\hat{\mathbf{p}}_i\in \mathbf{P}(\mathbf{S}_{k_i},\mathbf{E}_{l_i})}\exp(-\alpha \mathbf{D}_{\hat{\mathbf{p}}_i}(\mathbf{S}_{k_i},\mathrm{LoFA}(\mathbf{E}_{l_i}|\mathbf{w}))}_{\mathcal{L}_2(\mathbf{w})}\Big\}.
\label{eq:19}
\end{aligned}
\end{equation}
As we can see from Eq.~\ref{eq:19}, the M-step contains the optimization of two terms with respect to $\mathbf{w}$, \emph{i.e.,} $\mathcal{L}_1(\mathbf{w})$ and $\mathcal{L}_2(\mathbf{w})$. Specifically, the term of $\mathcal{L}_1(\mathbf{w})$ involves the minimization of the matching distance between source feature matrix $\mathbf{S}_{k_i}$ and the adapted emerging feature matrix $\mathrm{LoFA}(\mathbf{E}_{l_i}|\mathbf{w})$ along the optimal HiPa $\mathbf{p}^{(t)*}_i$ in the previous iteration step, while $\mathcal{L}_2(\mathbf{w})$ requires us to maximize their matching distance upon all possible HiPas including the optimal HiPa $\mathbf{p}^{(t)*}_i$. However, it is impractical to enumerate all HiPas in $\mathrm{P}(\mathbf{S}_{k_i},\mathbf{E}_{l_i})$. Similar to the above Hard-EM processing manner, a compromising strategy is to optimize $\mathcal{L}_1(\mathbf{w})$ and omit the influence of $\mathcal{L}_2(\mathbf{w})$, which also can be viewed as a ``hard'' operation. Actually, adjusting the value of the feature elements in $\mathrm{LoFA}(\mathbf{E}_{l_i}|\mathbf{w})$ to shorten the matching distance between it with $\mathbf{S}_{k_i}$ along one special HiPa congenitally increases the matching distance along the most other HiPas. If there are $n$ pairs of potentially matched $\mathbf{S}_{k_i}$ and $\mathbf{E}_{l_i}$, $i=1,\ldots n$, the final M-step can be formulated as
\begin{equation}
\mathbf{w}^{(t+1)}=\mathop{\arg\min}_\mathbf{w}\frac{1}{n}\sum_{i=1}^{n}\mathbf{D}_{\mathbf{p}^{(t)*}_i}(\mathbf{S}_{k_i},\mathrm{LoFA}(\mathbf{E}_{l_i}|\mathbf{w})),
\label{eq:20}
\end{equation} 
where $\mathbf{p}^{(t)*}_i$ is the optimal HiPa between $\mathbf{S}_{k_i}$ and $\mathbf{E}_{l_i}^{(t)}=\mathrm{LoFA}(\mathbf{E}_{l_i}|\mathbf{w}^{(t)})$, $i=1,\ldots,n$. The optimization strategy of Eq.~\ref{eq:20} has been detailed in Fig.~\ref{fig:LoFA}(a). Technically, the LoFA is achieved by a deep neural network, and the element-to-element distance loss is evaluated by MSE metric~\cite{lehmann2006theory}. Due to the complexity and the nonlinearity of the LoFA, we can not obtain the global optimal $\mathbf{w}^{(t+1)}$ by the closed derivation or the gradient descent methods. Therefore, we adopt the idea of Eq.~\ref{eq:10} and search for a relatively good $\mathbf{w}^{(t+1)}$, which meets the criterion $||\mathbf{w}^{(t+1)}-\mathbf{w}^{(t)}||\leq \epsilon$ as depicted in  Algorithm~\ref{algo:SrLFM}.

In brief, the ``Match'' step in the SLoMa algorithm is equivalent to the E-step as discussed above, which aims to match $\mathbf{S}_{k_i}$ and $\mathbf{E}^{(t)}_{l_i}$ and find their optimal HiPa $\mathbf{p}^{(t)*}_i$ using the DPW algorithm, while the ``Optimize'' step in SLoMa is the M-step, which seeks to retrain the $\mathrm{LoFA}(\cdot|\mathbf{w})$ to shorten the matching distance between $\mathbf{S}_{k_i}$ with $\mathrm{LoFA}(\mathbf{E}_{l_i}|\mathbf{w})$ along the previous optimal HiPa $\mathbf{p}^{(t)*}_i$ and then manufacture the current estimated $\mathbf{w}^{(t+1)}$.

\begin{figure*}[t!]
\centering
\includegraphics[width=1\linewidth]{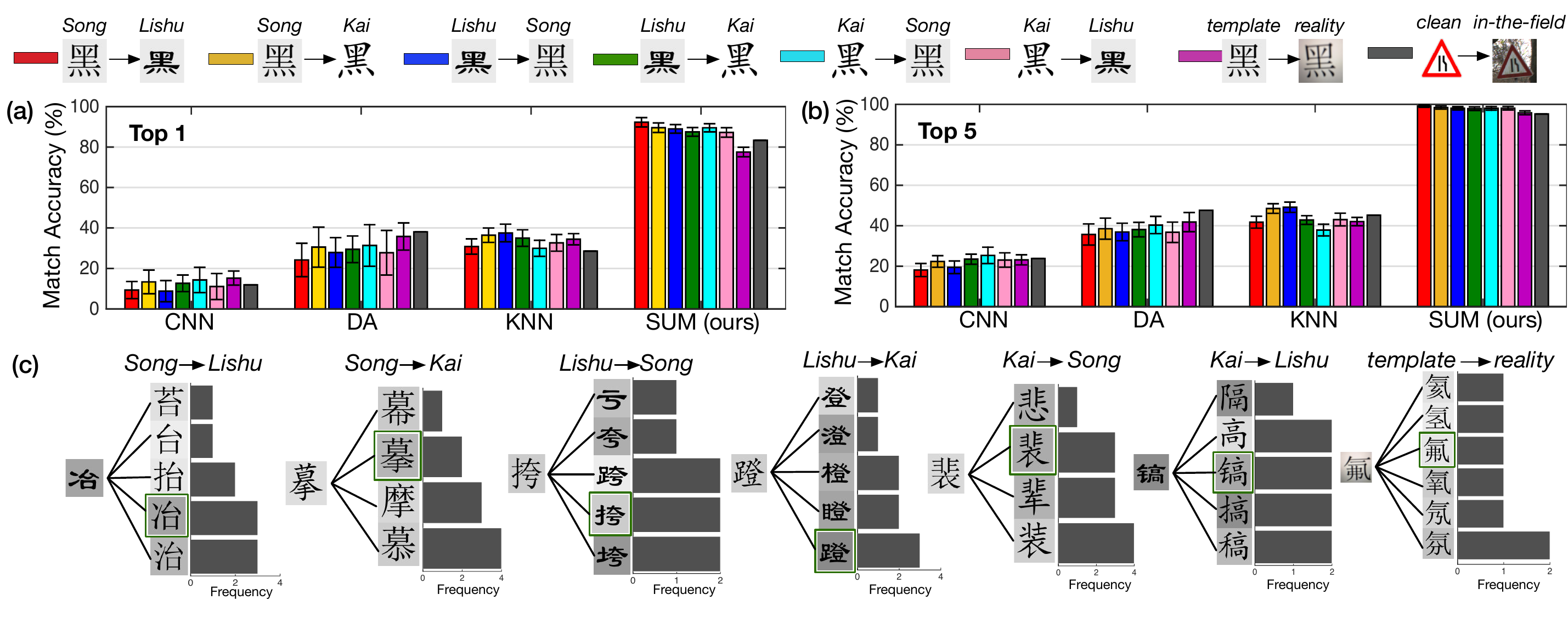}
\caption{(a) Top-1 and (b) top-5 match accuracy by different methods. Results of Chinese character matching (the front 7 colored bars for each approach) are reported using the averaged match accuracies with standard deviation (error bars) over 100 random experiments, while results of traffic sign matching (the last gray bar) are reported by only one experiment (since there are only 42 sign classes). (\emph{c}) Typical examples of cross-modality matching falsely with high frequency (by top 1 accuracy metric). In each subfigure of (\emph{c}), the emerging character to be matched is presented on left, and several of the most frequently matched characters in  seen modality inferred by SUM are presented on right. The bottom matching frequency is computed based on the statistics over 100 random experimental results. The image in green boxes denotes the ground-truth matched character.
}
\label{fig:ex1}
\end{figure*}

\subsection{Self-reinforcing Whole Image Matching}\label{sec:SWIM}
SLoMa operates based on the potentially matched subsets $\{\mathbf{S}_{k_i}\}_{i=1}^{n}$ and $\{\mathbf{E}_{l_i}\}_{i=1}^{n}$, whereas the difficult task setting we focus on 
supports no supervision for the ground-truth matched cross-modality congener objects. Thus, following the proposed self-learning process in Fig.~\ref{fig:human-self-learning}, we develop SWIM as described in Alg.~\ref{algo:SrWIM}. In each SWIM iteration $T$, the potentially matched $\mathbf{S}_{k_i}$ and $\mathbf{E}_{l_i}$ are explored by the statistics of DPW distances of all cross-modality data pairs and then absorbed into $\{\mathbf{S}_{k_i}\}_{i=1}^{n_T}$ and $\{\mathbf{E}_{l_i}\}_{i=1}^{n_T}$ to enable the inner SLoMa, where $k_i,l_i,n_T \in [1:N]$. Notably, with the evolution of the SWIM iteration $T$, the size of $\{\mathbf{S}_{k_i}\}_{i=1}^{n_T}$ and $\{\mathbf{E}_{l_i}\}_{i=1}^{n_T}$ progressively enlarges by adding $\alpha$ feature matrices per iteration until $\{\mathbf{E}_{l_i}\}_{i=1}^{n_T}$ covers all elements of $\{\mathbf{E}_{i}\}_{i=1}^{N}$ (i.e., $n_T=N$).

Fig.~\ref{fig:pipeline} illustrates the overall schematic of our SUM framework. In summary, we first train a CNN-based feature encoder on the seen modality by data augmentation. The feature encoder is then used to encode the images from the seen and emerging modality (i.e., $\{\mathbf{x}^s_i\}_{i=1}^N$ and $\{\mathbf{x}^e_i\}_{i=1}^N$) into feature matrices (i.e., $\{\mathbf{S}_i\}_{i=1}^{N}$ and $\{\mathbf{E}_i\}_{i=1}^{N}$). Finally, $\{\mathbf{S}_i\}_{i=1}^{N}$ and $\{\mathbf{E}_i\}_{i=1}^{N}$ are fed into SWIM to reveal their correspondence. Importantly, SUM requires supervision information neither on the cross-modality whole image matching nor on the cross-modality local feature matching, and it can be run fully autonomously using extremely few data.

\begin{figure*}[t!]
\centering
\includegraphics[width=0.95\linewidth]{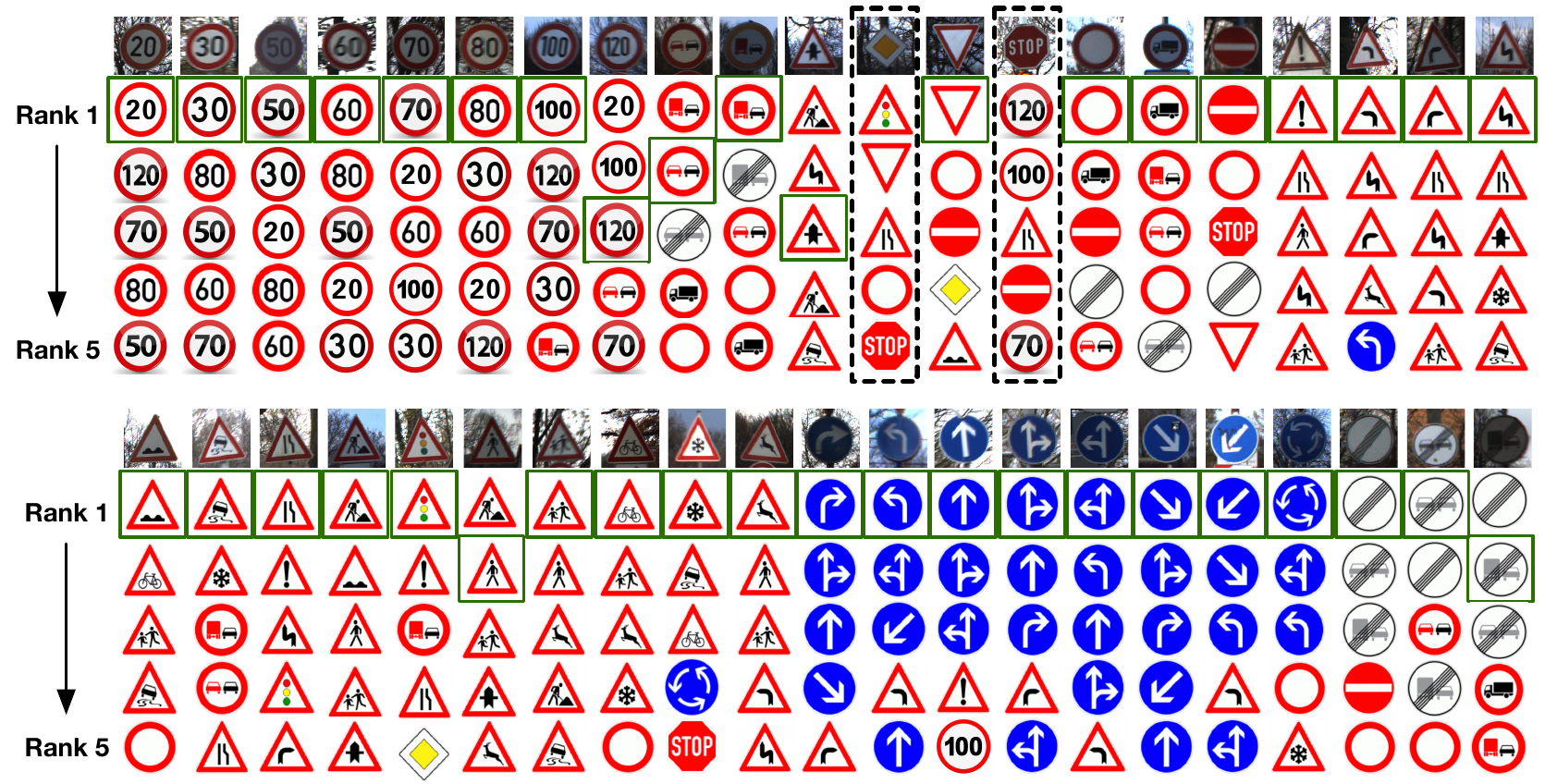}
\caption{Matching results for traffic signs. For one emerging sign, its matched templates ranking from 1 to 5 are presented. The image in green boxes denotes the ground-truth matched template.
}
\label{fig:traffic}
\end{figure*}

\section{Experiments}\label{exp}
\subsection{Matching Chinese Characters by One Template}
 Unlike most alphabet-based language characters, characters in Chinese, produced by using a stroke-based ideogram, have rich structural properties and diversely written fonts. The visual discrepancies between two Chinese characters in different fonts are often considerable despite being in the same character class. We collect one template image for each of the 3,755 frequently-used Chinese characters involving 3 different fonts, namely \emph{Song}, \emph{Lishu} and \emph{Kai}.  In addition, we also collect one realistic image for each of the 3,755 characters in \emph{Song} font. The designed tasks on Chinese characters include the following 7 cross-modality directions (seen$\rightarrow$emerging): \emph{Song}$\rightarrow$\emph{Lishu}, \emph{Song}$\rightarrow$\emph{Kai}, \emph{Lishu}$\rightarrow$\emph{Song}, \emph{Lishu}$\rightarrow$\emph{Kai}, \emph{Kai}$\rightarrow$\emph{Song}, \emph{Kai}$\rightarrow$\emph{Lishu} and \emph{Template Song}$\rightarrow$ \emph{Reality Song}. Notably, we construct an independent experiment by randomly selecting 100 different characters (i.e., $N=100$) from the 3,755 Chinese characters. The selected 100 character classes in two modalities share the same class space. Performance is evaluated by averaging the top-1 and top-5 match accuracies over 100 random experiments. Here, we compare SUM with other 3 alternative baselines, including CNN, domain adaptation (DA) and k-nearest neighbor (KNN) (see Appendix for more implementation details). The CNN, trained on the seen modality by data augmentation, is directly applied to classify the data in the emerging modality. This blunt baseline leads to a low matching performance regardless of the cross-modality directions, as shown in Fig.~\ref{fig:ex1}(a)-(b), confirming the conclusion that deep models struggle to naturally adapt to novel modality. Comparably, SUM achieves appealing averaged top-1 and top-5 match accuracies for the 7 directions, far exceeding those of DA and KNN. In addition, several typical examples of matching falsely with a high frequency by SUM w.r.t different task directions are displayed in Fig.~\ref{fig:ex1}(c). For one emerging character to be matched in each failure modality, we observe that the several most frequently matched characters in seen modality all share the analogous stroke unit or building structure. Coincidentally, these high-frequency matched characters in seen modality are deemed to be easily confused Chinese characters from the perspective of language~\cite{feldman1999semantic}. The experimental phenomena illustrates that SUM focuses on the characters' local structure to perform cross-modality  matching.

\subsection{Matching Traffic Signs by One Template}
In real scenes, traffic signs may be erected on a new outdoor site with different view angles, illumination conditions, background characteristics or chromatic aberrations from existing modalities. We select 42 classes of traffic signs from the German Traffic Sign Recognition Benchmark (GTSRB)~\cite{stallkamp2011german} as the research objects. Importantly, we collect one clean template for each sign class as the data in seen modality and single out 42 in-the-field images from GTSRB (one image per class) as data in emerging modality. The 42 emerging sign images populate a similar real background containing branches and sky (see Fig.~\ref{fig:traffic}). We also compare SUM with the CNN, DA and KNN via the match accuracy on the 42 traffic sign classes. As shown in Fig.~\ref{fig:ex1}(a)-(b), the significant advantages of SUM further demonstrates its effectiveness. Specifically, SUM achieves the top-1 match accuracy of $35/42$ and top-5 match accuracy of $40/42$. The detailed matching visualization results ranking from 1 to 5 for each emerging sign are depicted in Fig.~\ref{fig:traffic}. Intuitively, the top-5 most likely matched templates almost belong to the same semantic subclass of traffic signs possessing analogous geometric structures. For instance, the speed limit signs all contain a red ring encircling some black digits, while all warning signs contain a red triangle ring enfolding some graphic shapes. Two failure modalities are the signs of \emph{Priority} and \emph{Parking off} (marked by black dashed boxes), which are unusual among all traffic sign classes: excluding the \emph{Priority} and \emph{Parking off}, the rest 40 sign classes have their partners with more or less local similarities in structural appearance or components. Critically, the local similarities existing between two sign classes provide the cornerstone of cross-modality local feature adaptation (similar to the function of the basic strokes shared by different Chinese characters). In brief, these observations further support our argumentation that the backbone of SUM executing the whole image matching is the local feature matching.

\begin{figure*}[t!]
\centering
\includegraphics[width=1\linewidth]{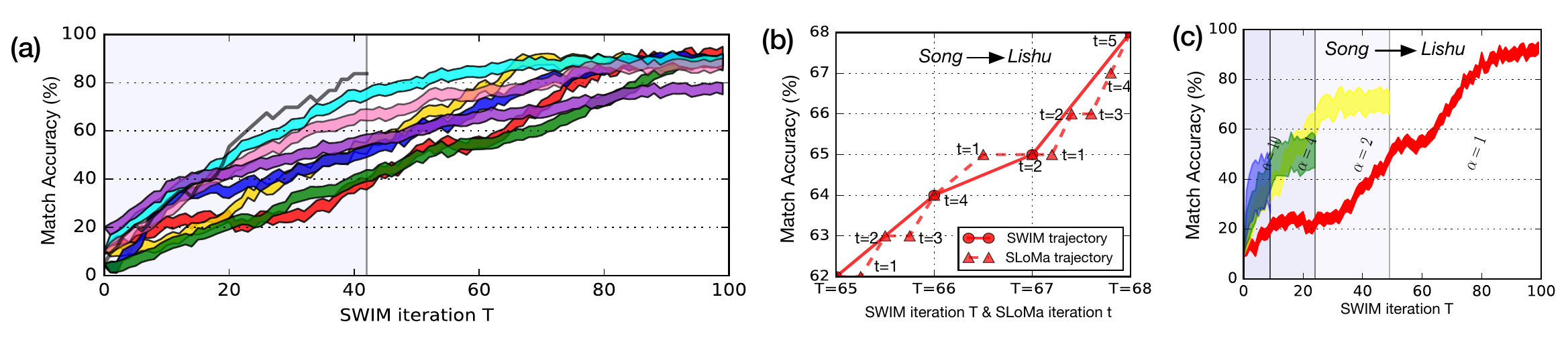}
\caption{
(a) SUM performance versus SWIM iteration $T$ for different task directions.
(b) Zoomed-in view when SWIM iteration $T$ evolves from 65 to 68 for \emph{Song}$\rightarrow$\emph{Lishu} task, wherein we also plot SUM performance versus SLoMa iteration $t$ (dashed line) inside each SWIM iteration. Here, we only show the result of a single experiment for clarity.
(c) SUM performance w.r.t the exploration step size $\alpha$ for \emph{Song}$\rightarrow$\emph{Lishu} task (four $\alpha$ settings are studied).
Note that (a) share the top-left legend of Fig.~\ref{fig:ex1}, and the color-filled curves in (a) and (c)  are bounded by the $\pm\mathrm{std}\%$ over 100 random experiments.
}
\label{fig:ex2}
\end{figure*}

\subsection{Improvement by Self-reinforcing Mechanism} 
For better comprehension of SUM, we also investigate the efficacy of self-reinforcing mechanism in both outer SWIM and inner SLoMa. As described in Algorithm~\ref{algo:SrLFM} and~\ref{algo:SrWIM}, we can obtain an updated LoFA after each SLoMa iteration $t$ as well as each SWIM iteration $T$. To track the immediate performance after different outer or inner self-reinforcing iterations, we successively utilize each immediate LoFA to adapt the $\{\mathbf{E}_{i}\}_{i=1}^N$ and then evaluate the match accuracy between these adapted substitutes and $\{\mathbf{S}_{i}\}_{i=1}^N$ (see Appendix). For SWIM, the matching performance of different visual tasks versus SWIM iteration $T$ are plotted in Fig.~\ref{fig:ex2}(a). We observe that at the beginning, the match accuracies are generally low (3--20$\%$) but climb steadily as the evolution of $T$ until the performance converges, highlighting the benefits of outer self-reinforcing iteration. Comparably, to reveal the capacity of SLoMa iteration $t$, we plot in Fig.~\ref{fig:ex2}(b) the matching performance on a random individual experiment of \emph{Song}$\rightarrow$\emph{Lishu} versus SLoMa iteration $t$ when SWIM iteration $T$ evolves from 65 to 68. Similarly, we see that the dashed line (i.e., performance versus SLoMa iteration) presents persistent growth inside each SWIM iteration ($T=$65--66, 66--67, and 67--68), illustrating the necessity of self-reinforcing iteration. Furthermore, we explore the effects of the exploration step size $\alpha$ in SWIM by setting different $\alpha$ values, as visualized in Fig.~\ref{fig:ex2}(c). Although a larger $\alpha$ results in a faster convergence speed of matching performance, the final match accuracy caused decline clearly relative to that by a smaller $\alpha$. This result can be reasonably explained by the intuition that it is more likely to absorb some mismatched pairs of feature matrices into the potentially matched cross-modality subsets when the exploration step is too large. Moreover, we can conclude from Fig.~\ref{fig:ex2}(c) that a larger $\alpha$ results in a greater matching performance oscillation (the standard deviation when $\alpha=10$ is the largest), but the one-by-one (i.e., $\alpha=1$) self-reinforcing exploration fashion creates a steady and progressive improvement on matching performance, which is analogous to the step-by-step self-learning process of human beings.

\section{Conclusion}
We propose an universal self-learning paradigm at cognition level to encourage the cross-modality data matching based on extremely scarce templates. Inspiringly, a sophisticated self-reinforcing unsupervised matching framework for the autonomous annotation of data in an emerging modality is designed. Our framework is empirically proven to be effective on revealing the cross-modality correspondence between 2D structure-preserving image objects. Despite not achieving one hundred percent, the high matching performance for data in one emerging modality surpasses other feasible approaches, considerably reducing the manual labeling cost and moving towards continual machine learning. It is promising to make our framework compatible with several hybrid emerging modalities instead of only one in each cross-modality match manipulation, which is left for future work.


%

\appendices
\section{Proof of the Optimality of $\mathbf{p}^*$}
Considering two 2D feature matrices, $\mathbf{S}=[\mathbf{s}_{ij}]\in \mathbb{R}^{H^s\times W^s}$ and $\mathbf{E}=[\mathbf{e}_{ij}]\in \mathbb{R}^{H^e\times W^e}$, we can calculate their DPW distance by Eq.~\ref{eq:DPW_distance_solve} and the corresponding optimal HiPa $\mathbf{p}^*$ by Algorithm~\ref{algo:oHWP}. In this section, we provide the theoretical proof for the optimality of $\mathbf{p}^*$.

First, we prove the optimality of first-level path nodes. If $H^s$$=$$1$, there is only one feasible case for first-level warping path between $\mathbf{S}$ and $\mathbf{E}$, i.e., $\{(1,1),(1,2),\ldots,(1,H^e)\}$ , whose corresponding DPW distance naturally is $\sum_{i=1}^{H^e}\mathcal{DTW}(\mathbf{S}(1),\mathbf{E}(i))$. Similarly, if $H^e=1$, there also is only one feasible case for the first-level path nodes, i.e., $\{(1,1),(2,1),\ldots,(H^s,1)\}$, whose corresponding DPW distance is $\sum_{i=1}^{H^s}(\mathbf{S}(i),\mathbf{E}(1))$. When $H^s>1$ and $H^e>1$, we let $\{p^{(1)}_1,\ldots,p^{(1)}_{C}\}$ is the optimal first-level warping path node set between $\mathbf{S}(1$$:$$\ h^s)$ and $\mathbf{E}(1$$:$$\ h^e)$, $h^s\in [1:H^s]$ and $h^e\in[1:H^e]$ (Note that $\mathbf{S}(1$$:$$\ h^s)\in \mathbb{R}^{h^s\times W^s}$ represents a feature sub-matrix, and similar for $\mathbf{E}(1: h^e)\in \mathbb{R}^{h^e\times W^e}$). According to HBC, we have $p^{(1)}_C=(h^s,h^e)$. Suppose $p^{(1)}_{C-1}:=(a,b)$, then $(a,b)\in\{(h^s-1,h^e-1),(h^s-1,h^e),(h^s,h^e-1)\}$ because of the HSC; thus, $\{p^{(1)}_1,\ldots,p^{(1)}_{C-1}\}$ must be the optimal first-level warping path node set between $\mathbf{S}(1:a)$ and $\mathbf{E}(1:b)$. Because that if it is not the optimal, there must be a better first-level warping path node set $\{\hat{p}^{(1)}_1,\ldots,\hat{p}^{(1)}_M\}$ satisfying:
\begin{equation}
\begin{aligned}
&\mathbf{D}_{\{\hat{p}^{(1)}_1,\ldots,\hat{p}^{(1)}_M\}}(\mathbf{S}(1:a),\mathbf{E}(1:b))\\
<&\mathbf{D}_{\{p^{(1)}_1,\ldots,p^{(1)}_{C-1}\}}(\mathbf{S}(1:a),\mathbf{E}(1:b)), 
\end{aligned}
\end{equation}
then the total matching distance between $\mathbf{S}(1:h^s)$ and $\mathbf{E}(1:h^e)$ along $\{p^{(1)}_1,\ldots,p^{(1)}_C\}$ will fall into the following fact:
\begin{equation}
\begin{aligned}
&\ \mathbf{D}_{\{\hat{p}^{(1)}_1,\ldots,\hat{p}^{(1)}_M,p^{(1)}_C\}}(\mathbf{S}(1:h^s),\mathbf{E}(1:h^e))\\
=&\ \mathbf{D}_{\{\hat{p}^{(1)}_1,\ldots,\hat{p}^{(1)}_M\}}(\mathbf{S}(1:a),\mathbf{E}(1:b))+\mathcal{DTW}(\mathbf{S}(h^s),\mathbf{S}(h^e))\\
<&\ \mathbf{D}_{\{p^{(1)}_1,\ldots,p^{(1)}_C\}}(\mathbf{S}(1:h^s),\mathbf{E}(1:h^e))\\
=&\ \mathbf{D}_{\{p^{(1)}_1,\ldots,p^{(1)}_{C-1}\}}(\mathbf{S}(1:a),\mathbf{E}(1:b))+\mathcal{DTW}(\mathbf{S}(h^s),\mathbf{E}(h^e)),
\end{aligned}
\end{equation}
which means that $\{\hat{p}^{(1)}_1,\ldots,\hat{p}^{(1)}_M,p^{(1)}_C\}$ is a better first-level path node set than $\{p^{(1)},\ldots,p^{(1)}_C\}$. This conclusion is in contradiction with the hypothesis that $\{p^{(1)}_1,\ldots,p^{(1)}_C\}$ is the optimal first-level path between $\mathbf{S}(1:h^s)$ and $\mathbf{E}(1:h^e)$; thus, 
\begin{equation}
\begin{aligned}
\mathcal{D}(a,b) &= \mathcal{DPW}(\mathbf{S}(1:a),\mathbf{E}(1:b))\\
&=\mathbf{D}_{\{p^{(1)}_1,\ldots,p^{(1)}_{C-1}\}}(\mathbf{S}(1:a),\mathbf{E}(1:b)),
\end{aligned}
\end{equation}
where $\mathcal{D}\in \mathbb{R}^{H^s\times H^e}$ is the hierarchical accumulated distance matrix. Furthermore,  
\begin{equation}
\begin{aligned}
&\mathcal{D}(h^s,h^e)= \mathcal{DPW}(\mathbf{S}(1:h^s),\mathbf{E}(1:h^e))\\
=&\ \mathbf{D}_{\{p^{(1)}_1,\ldots,p^{(1)}_C\}}(\mathbf{S}(1:h^s),\mathbf{E}(1:h^e))\\
=&\ \mathbf{D}_{\{p^{(1)}_1,\ldots,p^{(1)}_{C-1}\}}(\mathbf{S}(1:a),\mathbf{E}(1:b))+\mathcal{DTW}(\mathbf{S}(h^s),\mathbf{E}(h^e))\\
=&\ \mathcal{D}(a,b)+\mathcal{DTW}(\mathbf{S}(h^s),\mathbf{E}(h^e))\\
=&\ \mathop{\min}\{\mathcal{D}(h^s-1,h^e-1),\mathcal{D}(h^s-1,h^e),\mathcal{D}(h^s,h^e-1)\}+\\
&\ \mathcal{DTW}(\mathbf{S}(h^s),\mathbf{E}(h^e)),
\end{aligned}
\end{equation}
which keeps the same with dynamic programming algorithm in Eq.~\ref{eq:DPW_distance_solve}. In other words, the calculation for DPW distance and optimal HiPa based on Eq.~\ref{eq:DPW_distance_solve} and~\ref{eq:first-level nodes} guarantees the optimality of its first-level path node set $\{p^{(1)*}_1,\ldots,p^{(1)*}_L\}$. 

Next, we prove the optimality of the second-level path node set. Without loss of generality, we present the proof only on $\{p^{(2)}_1,\ldots,p^{(2)}_{c_1}\}|  p^{(1)}_{c_1}$ (other second-level nodes can be proved in the same manner). 
Obviously, this second-level path node set defines an alignment between $\mathbf{S}(1)$ and $\mathbf{E}(1)$ because $p^{(1)}_{c_1}=(1,1)$, where $\mathbf{S}(1)$ and $\mathbf{E}(1)$ are the first row feature sequence of $\mathbf{S}$ and $\mathbf{E}$, respectively. If $W^s$$=$$1$, there is only a feasible second-level path node set $\{(1,1),(1,2),\ldots,(1,W^e)\}$ between $\mathbf{S}(1)$ and $\mathbf{E}(1)$, whose corresponding DTW distance is $\mathcal{DTW}(\mathbf{S}(1),\mathbf{E}(1))=\sum_{i=1}^{W^e}\mathbf{d}(\mathbf{S}(1,1),\mathbf{E}(1,i))$, where $\mathbf{S}(1,1)=\mathbf{s}_{11}$ and similar for $\mathbf{E}$. If $W^e$$=$$1$, there is also only a feasible second-level path node set $\{(1,1),(2,1),\ldots,(W^s,1)\}$ between $\mathbf{S}(1)$ and $\mathbf{E}(1)$, whose corresponding DTW distance is $\mathcal{DTW}(\mathbf{S}(1),\mathbf{E}(1))=\sum_{i=1}^{W^s}\mathbf{d}(\mathbf{S}(1,i),\mathbf{E}(1,1))$. When $W^s$$>$$1$ and $W^e$$>$$1$, we let $\{p^{(2)}_1,\ldots,p^{(2)}_{C}\}$ be the optimal second-level path node set between $\mathbf{S}(1,1$$:$$w^s)$ and $\mathbf{E}(1,1$$:$$w^e)$, $w^s$$\in$$[1$$:$$W^s]$ and $w^e$$\in$$ [1$$:$$W^e]$ (Note that $\mathbf{S}(1,1$$:$$w^s)\in \mathbb{R}^{w^s}$ represent a feature sub-sequence, and similar for $\mathbf{E}(1,1$$:$$w^e)\in \mathbb{R}^{w^e}$). The HBC implies $p^{(2)}_C=(w^s, w^e)$. Let $p^{(2)}_{C-1}:=(c,d)$, then HSC implies $(c,d)\in\{(w^s-1,w^e-1),(w^s-1,w^e),(w^s,w^e-1)\}$. Similarly, we can conclude that $\{p^{(2)}_1,\ldots,p^{(2)}_{C-1}\}$ is the optimal second-level path node set between sub-sequence $\mathbf{S}(1,1$$:$$c)$ and $\mathbf{E}(1,1$$:$$d)$; otherwise, there should be a better second-level warping path $\{\hat{p}^{(2)}_1,...,\hat{p}^{(2)}_M\}$ satisfying:
\begin{equation}
\begin{aligned}
&\ \mathbf{D}_{\{\hat{p}^{(2)}_1,...,\hat{p}^{(2)}_M\}}(\mathbf{S}(1,1:c),\mathbf{E}(1:1:d)) \\
<&\ \mathbf{D}_{\{p^{(2)}_1,\ldots,p^{(2)}_{C-1}\}}(\mathbf{S}(1,1:c),\mathbf{E}(1:1:d)), 
\end{aligned}
\end{equation}
where the notation $\mathbf{D}_{\{p^{(2)}_1,\ldots,p^{(2)}_{c}\}}(\mathbf{x},\mathbf{y})$ denotes the matching distance between two sequence feature $\mathbf{x}$ and $\mathbf{y}$ along the second-level path node set $\{p^{(2)}_1,\ldots,p^{(2)}_{c}\}$. Consequently, we can get a undesirable case:
\begin{equation}
\begin{aligned}
&\mathbf{D}_{\{\hat{p}^{(2)}_1,\ldots,\hat{p}^{(2)}_M,p^{(2)}_C\}}(\mathbf{S}(1,1:w^s),\mathbf{E}(1,1:w^e))\\
=&\ \mathbf{D}_{\{\hat{p}^{(2)}_1,\ldots,\hat{p}^{(2)}_M\}}(\mathbf{S}(1,1:c),\mathbf{E}(1,1:d))+\mathbf{d}(\mathbf{s}_{1w^s},\mathbf{e}_{1w^e})\\
<&\ \mathbf{D}_{\{p^{(2)}_1,\ldots,p^{(2)}_C\}}(\mathbf{S}(1,1:w^s),\mathbf{E}(1,1:w^e))\\
=&\ \mathbf{D}_{\{p^{(2)}_1,\ldots,p^{(2)}_{C-1}\}}(\mathbf{S}(1,1:c),\mathbf{E}(1,1:d))+\mathbf{d}(\mathbf{s}_{1w^s},\mathbf{e}_{1w^e}),
\end{aligned}
\end{equation}
which conflicts with the hypothesis that $\{p^{(2)}_1,\ldots,p^{(2)}_C\}$ is the optimal second-level  path node set between $\mathbf{S}(1,1:w^s)$ and $\mathbf{E}(1,1:w^e)$. Thus,
\begin{equation}
\begin{aligned}
&\ \mathcal{DTW}(\mathbf{S}(1,1:c),\mathbf{E}(1,1:d))\\
=&\ \mathbf{D}_{\{p^{(2)}_1,\ldots,p^{(2)}_{C-1}\}}(\mathbf{S}(1,1:c),\mathbf{E}(1,1:d)),
\end{aligned}
\end{equation}
and
\begin{equation}
\begin{aligned}
&\mathcal{DTW}(\mathbf{S}(1,1:w^s),\mathbf{E}(1,1:w^e))\\
=&\ \mathbf{D}_{\{p^{(2)}_1,\ldots,p^{(2)}_K\}}(\mathbf{S}(1,1:w^s),\mathbf{E}(1,1:w^e))\\
=&\ \mathcal{DTW}(\mathbf{S}(1,1:c),\mathbf{E}(1,1:d))+\mathbf{d}(\mathbf{s}_{1w^s},\mathbf{e}_{1w^e})\\
=&\ \mathop{\min}\{\mathcal{DTW}(\mathbf{S}(1,1:w^s-1),\mathbf{E}(1,1:w^e-1)),\\
&\qquad \ \ \mathcal{DTW}(\mathbf{S}(1,1:w^s-1),\mathbf{E}(1:1:w^e)),\\
   &\qquad\ \ \mathcal{DTW}(\mathbf{S}(1,1:w^s),\mathbf{E}(1:w^e-1))\}+\\
   &\qquad\ \ \mathbf{d}(\mathbf{s}_{1w^s},\mathbf{e}_{1w^e}).
\end{aligned}
\end{equation}
The recursive arithmetic above is actually the DTW dynamic programming algorithm. Therefore, the second-level path node set $\{p^{(2)*}_k|\ p^{(1)*}_l, k=1,\ldots,c_l, l=1,\ldots,L\}$ solved by Eq.~\ref{eq:second-level nodes} is optimal among all feasible second-level paths.

\section{Experiment Details}
\subsection*{Experimental Data}
The SUM framework is validated on two types of experimental data: Chinese characters and traffic signs. For Chinese characters, we focus on the 3755 frequently-used Chinese characters and collect one template image for each character involving three different written fonts (\emph{Song, Lishu and Kai}) and one realistic shooting case (\emph{Song}). Hence, there are $3755\times4$ collected Chinese character images. Chinese is universally accepted to be one difficult language character all over the world since its sophisticated stroke structure and various written fonts. In order to shorten the experimental period and get more comprehensive conclusions rapidly, we design 100 experiments on Chinese characters each of which is designed into a 100-way cross-font matching task.  During each individual experiment, the 100 task character classes are selected from the 3755 frequently-used characters at random. 
Another dataset entails the traffic signs, which are built through our manual collection for the clean exemplars and the in-the-field traffic sign dataset, German Traffic Sign Recognition Benchmark (GTSRB)~\cite{stallkamp2011german}. The GTSRB contains 43 different traffic sign classes, and most of them has been presented in Fig.~\ref{fig:traffic}. Excluding the sign class of ``\emph{Speed (80) Limit Cancel}'', the remaining 42 sign classes are all chosen as the experimental subjects. The reason that we omit this sign class comes from two aspects. (1) It is hard for us to collect one clean exemplar for this sign class. (2) There are no satisfactory in-the-field image for this sign class in the GTSRB (populating a similar real background with other 42 signs containing the branches and the sky). We single out one image from the GTSRB for each of the 42 sign classes by following the fundamental principle that they must populate a similar environment. This principle aims at simulating the case that the agent is required to adapt to a set of emerging traffic signs erected on a new outdoor site. 

\subsection*{Computational Models}
The feature encoder in the SUM framework is built by extracting the convolutional blocks from the modality-specifical CNN which is well trained to classify the data in the seen modality. Clearly, the model architectures of modality-specifical CNN with regard to the Chinese character matching and the traffic sign matching are detailed in Table~\ref{table:model_architecture}. Given a set of template character images or traffic sign images (these images populate one same modality, i.e., seen modality), we first augment it by performing the translation, shearing and scaling transformation on raw template images so that the number of the training samples per class reach a sizable scale (100 data per class). Then, the augment dataset of the seen modality are utilized to train the CNN by cross-entropy classification loss and stochastic gradient-descent back propagation algorithm NADAM~\cite{dozat2016incorporating} (an enhanced version of ADAM~\cite{kingma2014adam}) with the initial learning rate $1\times10^{-5}$ and the schedule decay $0.004$ until convergence. Especially, the nonlinearity units for all weight layers are ReLUs~\cite{nair2010rectified} except those of final convolutional layer are Sigmoids. We do this because that we hope the extracted feature matrices hold richer $0-1$ continuous value information (our preliminary experimental results indicate that this kind of output benefits to the eventual matching performance). The first three blocks of the well-trained CNN are fixed as the feature encoder, which takes the Chinese character images or the traffic sign images as input and output their corresponding feature matrices. In our experiments, the size of all images has been processed into the RGB form with size of $80\times80\times3$; thus the size of feature matrix is $10\times10\times160$ for the Chinese character tasks and $10\times10\times80$ for the traffic sign tasks, respectively. Naturally, the feature encoder is specific for the seen modality, which struggles to extract distinguishing and significant patterns or features for the incoming data of an emerging modality. 

\begin{table*}[t!]
\begin{center}
\scalebox{0.9}{
\begin{tabular}{lllllll}
\hline
 & Block 1 & Block 2 & Block 3 & Block 4 & Block 5 & Block 6  \\
\hline
\multirow{4}{*}{CNN for Chinese character tasks}
 & conv-$3\times3\times40$ & conv-$3\times3\times80$ & conv-$3\times3\times160$ & fc-$2048$ & fc-$2048$ &softmax-$3755$ \\
 &conv-$3\times3\times40$ & conv-$3\times3\times80$ &conv-$3\times3\times160$ \\
 & stride-$1$, pad-$1$&  stride-$1$, pad-$1$ &  stride-$1$, pad-$1$ \\
 & maxpooling-$2\times2$ & maxpooling-$2\times2$ & maxpooling-$2\times2$ \\
\hline
\multirow{4}{*}{CNN for traffic sign tasks}
 & conv-$3\times3\times20$ & conv-$3\times3\times40$ & conv-$3\times3\times80$ & fc-$50$ & fc-$50$ &softmax-$42$ \\
 &conv-$3\times3\times20$ & conv-$3\times3\times40$ &conv-$3\times3\times80$ \\
 & stride-$1$, pad-$1$&  stride-$1$, pad-$1$ &  stride-$1$, pad-$1$ \\
 & maxpooling-$2\times2$ & maxpooling-$2\times2$ & maxpooling-$2\times2$ \\
\hline
\end{tabular}}
\end{center}
\caption{Detailed architectures of the CNN trained on the data of the seen modalities. The ‘‘stride” is the convolution stride and the ‘‘pad” is the spatial padding on feature maps.}
\label{table:model_architecture}
\end{table*}

The LoFA aims at executing element-to-element conversion from an emerging modality to an seen modality. A feature element is actually a feature vector of $C$-dimension. In our experiments, we choose multi-layer perceptron (MLP)~\cite{bishop1995neural} as the LoFA. Certainly, more sophisticated networks can replace it, like recurrent neural networks (RNN) or long short-term memory network (LSTM)~\cite{hochreiter1997long}, auto-encoder~\cite{hinton2006reducing}. Specifically, each layer of MLP is followed by a sigmoid nonlinear activation function and a dropout operation ($p=0.2$). We adopt the 2-layer MLP for the two tasks. For the Chinese character matching tasks, the input size and the output size are all $160$ and each hidden layer disposes $400$ neuron nodes. For the traffic sign matching tasks, the input size and the output size are all $80$ and each hidden layer disposes $200$ neuron nodes. The training of LoFA is done by using NADAM optimization algorithm with the initial learning rate $1\times10^{-3}$ and the schedule decay $0.004$. As shown in Fig.~\ref{fig:LoFA}(a), the LoFA is optimized based on many pairs of $(\mathbf{e}_{h^ew^e}, \mathbf{s}_{h^sw^s})$ wherein $\mathbf{e}_{h^ew^e}$ is determined to match $\mathbf{s}_{h^sw^s}$. In this context, the training process can be viewed as a regression process where the input is  $\mathbf{e}_{h^ew^e}$ and its output is expected to approach $\mathbf{s}_{h^sw^s}$. The LoFA needs to be trained during each SLoMa iteration, and this operation ceases until the discrepancy between the weights of two adjacent iterative steps is small enough. In our experiments, the iteration error threshold $\epsilon$ is set as $1\times10^{-3}$.

\subsection*{Track the Immediate Matching Performance}\label{sec:Track the Immediate Matching Performance}
In main text, we explore the effectiveness of self-reinforcing mechanisms of the SUM framework by plotting the immediate matching performance of each SLoMa iteration or each SWIM iteration. In this section, we introduce how to track the immediate maching performance under the SUM framework. As described by Algorithm~\ref{algo:SrLFM} and~\ref{algo:SrWIM}, after each SLoMa iteration $t$ or after each SWIM iteration $T$, we can get a updated $\mathrm{LoFA}(\cdot|\hat{\mathbf{w}})$. Using this LoFA, we can obtain the new adapted feature matrices in the emerging modality:
\begin{equation}
\hat{\mathbf{E}}_i = \mathrm{LoFA}(\mathbf{E}_i|\hat{\mathbf{w}}),\ i=1,\ldots,N.
\end{equation}
The final cross-modality match between $\mathbf{E}$ and $\mathbf{S}$ can operate in the following manner:
\begin{equation}
\mathbf{E}_i \leftrightarrow \mathop{\arg\min}_{\mathbf{S}\in\{\mathbf{S}_{i}\}_{i=1}^N}\mathcal{DPW}(\mathbf{S},\hat{\mathbf{E}_i}),\ i =1,\ldots,N, 
\label{eq:22}
\end{equation}
where the symbol ``$x \leftrightarrow$ y'' represents that $x$ is determined to match $y$. Thus, we can assess the match accuracy after each SLoMa iteration and each SWIM iteration.

\subsection*{Comparison Baselines}
\noindent \textbf{Convolutional Neural Networks (CNN)}.
The first comparison baseline is straightforward: using the well-trained CNN on the seen modality as the classifier for the emerging modality, i.e., the feature encoder plus the last 3 blocks (Block 4, Block 5 and Block 6). In other words, the CNN which has been manufactured on the seen modality is directly applied to the classification for the data in the emerging modality. For Chinese character matching task, each individual experiment is a 100-way classification task, whereas the pre-trained CNN is originally designed for 3755 ways. Therefore, we omit the irrelevant links  between the penultimate block (namely Block 5) and softmax block (namely Block 6), and only retain the 100 output channels of task classes during each experiment. This baselines aims at illustrating the drawback or the incapability of the deep models when dealing with the unseen modality. 

\noindent \textbf{Domain Adaptation (DA)}.
DA is a hot research topic focusing on eliminating the domain shifts or the dataset biases to adapt the model which is learned on a source domain to a target domain~\cite{gopalan2011domain,tzeng2014deep}. Generally, there are some supervised samples in source domain and some unsupervised samples to be classified in target domain. The goal of DA is to develop a transferable classifer to cope with the data in target domain. For our tasks, the seen modality can be regarded as the source domain and the emerging modality is the target domain. We adopt the classical and state-of-the-art DA method, \emph{Deep Domain Confusion}~\cite{tzeng2014deep}, as the comparison baseline. In detail, this method introduces an additional block (called \emph{Block Adapt} here) and a \emph{Maximum Mean Discrepancy} (MMD) loss function to learn the domain invariant representation.
In fact, MMD is a nonparametric distance metric to evaluate the similarity between two different distributions in Reproducing Kernel Hilbert Spaces $\mathcal{H}$ (RKHS) with the feature mapping function $f(\cdot|\mathbf{w})$, which is formalized as:
\begin{equation}
\mathrm{MMD}(P_s, P_t)  = \big\|\mathrm{E}_{\mathbf{x}^s\sim P_s}(f(\mathbf{x}^s|\mathbf{w}))-\mathrm{E}_{\mathbf{x}^t\sim P_t}(f(\mathbf{x}^t|\mathbf{w}))\big\|_{\mathcal{H}}^{2},
\end{equation}
where $P_s$ and $P_t$ denote the distributions of the source domain and the target domain, respectively. Different from the above CNN baseline, we do data augmentation (translation, shearing and scaling transformation) on both the seen modality and the emerging modality. So we have sufficient supervised data in the seen modality and sufficient unsupervised data in the emerging modality. Based on the same CNN architecture with previous baseline, we insert the Block Adpat between the Block 5 and the Block 6. Actually, the Block 5 is a fully connected (FC) layer and the Block 6 is a 3755-way softmax layer. The Block Adapt is a FC layer with $d$ neurons and ReLU nonlinear function ($d=512$ for the Chinese character tasks and $d=32$ for the traffic sign tasks). Let the output of Block Adapt to be $\phi(\cdot)$,  then an empirical approximation to the MMD distance is computed as follows:
\begin{equation}
\mathrm{MMD}(X^s, X^e)  = \Big|\Big|\frac{1}{|X^s|}\sum_{\mathbf{x}^s\in X^s}\phi(\mathbf{x}^s)-\frac{1}{|X^e|}\sum_{\mathbf{x}^e\in X^e}\phi(\mathbf{x}^e)\Big|\Big|,
\end{equation}
where $X^s$ denotes the augmented sample set in the seen modality and $X^e$ the augmented sample set in the emerging modality. The MMD distance reflects the distribution discrepancy with respect to data representations $\phi(\cdot)$ from two different modalities and seeks to confuse the domain shifts by learning a domain-invariant representaion $\phi(\cdot)$. Meanwhile, we hope that the data representations $\phi(\cdot)$ can lead to accurate classification in the seen modality; thus, we combine the above two criteria into the following loss:
\begin{equation}
\mathcal{L}_{DA} = \mathcal{L}_{cla}(X^s, Y^s) + \lambda\mathrm{MMD}^2(X^s, X^e),
\label{eq:25}
\end{equation}
where the first term $\mathcal{L}_{cla}(X^s, Y^s)$ is the cross-entropy classification loss formed by data in the seen modality ($Y^s$ is the label set), and $\lambda$ is a trade-off hyperparameter determining the extent of the domain confusion. During the practical training process, the 
objective function in Eq.~\ref{eq:25} is performed in the mini-batch manner. In other words, $X^s$ or $X^e$ is a mini-batch of data sampled from their corresponding whole augmented datasets. Note that we do not train the architecture from scratch; instead we use the pre-trained Block 1-6 prior to the feature encoder construction as our start point. The $\lambda$ is set to $0.1$, which is suggested by our preliminary experiments. The placement and the layer dimension of the Block Adapt is chosen according to the conclusion in~\cite{tzeng2014deep}, which claims that the MMD distance can be used for model selection. The overall DA model is trained using the NADAM algorithm with the initial learning rate $1\times10^{-3}$ and the size of mini-batch 128. Once well trained, the DA model (i.e., Block 1-5 + Block Adapt + Block 6) is ready to classify the data in the emerging modality. Similar to the CNN baseline, for the Chinese character matching tasks, we close the non-task channels in the softmax layer (Block 6) during each random 100-way experiment and only compute the probability over these 100 task classes. 

\noindent \textbf{K-Nearest Neighbor (KNN).}
To verify the rationality of using DPW to evaluate the similarities between the feature matrices in the seen and emerging modalities, we design the third comparison baseline, KNN. Inside the KNN, most of the procedures and operations keep the same with those of the SUM framework except of the last matching step. Specifically, in SUM, we perform the DPW distance based nearest nerghbor matching (refer to Eq.~\ref{eq:22}), but in KNN baseline we use the naive point-wise distance to make final match decision:
\begin{equation}
\begin{aligned}
&\mathbf{E}^{*}_i = \mathrm{LoFA}(\mathbf{E}_i|\mathbf{w}^{*}),\ i=1,\ldots,N,\\
&\mathbf{E}_i \leftrightarrow \mathop{\arg\min}_{\mathbf{S}\in\{\mathbf{S}_{i}\}_{i=1}^N}\big|\big|\mathbf{S}-\mathbf{E}^*_i\big|\big|_1,\ i =1,\ldots,N,
\label{eq:27} 
\end{aligned}
\end{equation}
where $\mathrm{LoFA}(\cdot|\mathbf{w}^*)$ is the ultimate updated LoFA, and $||A||_1=\sum_{i}\sum_{j}|a_{ij}|$. Apparently, this baseline do not consider the factor of the position matching, and its poor matching performance compared with the  SUM framework corroborates from the side the conclusion that the SUM realizes the whole image matching by aligning the local features.

\subsection*{Top-1 and Top-5 Match Accuracy}
In our experiments, we report the matching performance by the metric of the top-1 or top-5 match accuracy. The metric of top-1 match is straightforward: for each of the images in the emerging modality, if its best matching result in the seen modality inferred by one matching algorithm certainly belongs to the same class, we view this case as a ``correct match''; otherwise, it is an ``incorrect match ''. For the SUM framework, the output result of the SWIM algorithm is naturally the top-1 matching result. For the CNN baseline and the DA baseline, we use the class label whose corresponding channel in the Block 6 possesses the largest predicted probability as the ultimate matching result. For the KNN baseline, we get the best matching result by adopting the calculation in Eq.~\ref{eq:27}. What needs to be explained carefully is the metric of top-5 match. In this case, for one data in the emerging modality, we compute its 5 best matching alternatives in the seen modality. If the ground-truth matching result exists among the 5 alternatives, we view this as a ``correct match''; otherwise, it is an ``incorrect match''. For the SUM framework, we list the 5 best matching results in the seen modality for each data in the emerging modality by ranking their DPW distances from small to large (the rank about DPW distance can be obtained by Eq.~\ref{eq:22}). For the KNN baseline, analogously, we can also obtain the 5 alternatives by ranking the point-wise distance $\big|\big| \mathbf{S}_j-\mathbf{E}^*_i\big|\big|_1,\ j=1,\ldots,N$ and taking the first 5 smallest ones. For the CNN baseline and the DA baseline, we take as the matching results the 5 class labels who respectively possess the 5 largest predicted probabilities among all channels in the Block 6. Regardless of which method, the ultimate top-1 or top-5 match accuracy is computed as follow:
\begin{equation}
\mathrm{Acc}=\frac{\mathrm{Num} (correct\ match\ case) }{\mathrm{Num}(data\ in\ emerging\ modality)}\times 100\%.
\end{equation}

\ifCLASSOPTIONcompsoc
  \section*{Acknowledgments}
\else
  \section*{Acknowledgment}
\fi

This work was supported in part by the National Natural Science Foundation of China under Grant 61473167 and Grant 61673394, in part by the Beijing Natural Science Foundation under Grant L172037, and in part by the German Research Foundation through Project Cross Modal Learning under Grant NSFC 61621136008/DFG TRR-169.

\ifCLASSOPTIONcaptionsoff
  \newpage
\fi



%
\bibliographystyle{IEEEtran}
\bibliography{ref}

%
\vspace{-1em}
\begin{IEEEbiography}[{\includegraphics[width=1in,height=1.25in,clip,keepaspectratio]{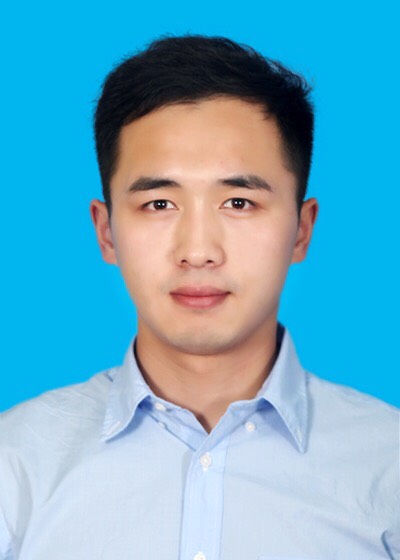}}]{Jiang Lu}
received the B.S. degree from Tsinghua University, Beijing, China, in 2013, where he is currently working toward the Ph.D. degree in the Department of Automation.
His research interests include machine learning, deep learning and computer vision.
\end{IEEEbiography}
\vspace{-1em}

\begin{IEEEbiography}[{\includegraphics[width=1in,height=1.25in,clip,keepaspectratio]{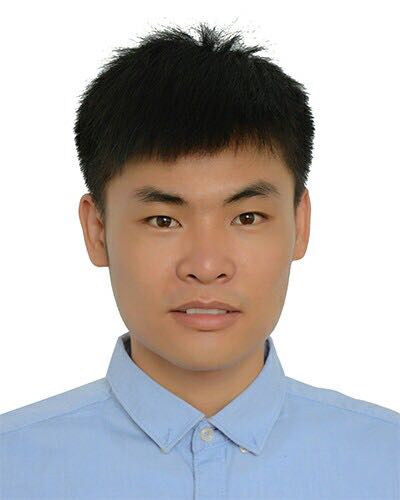}}]{Lei Li}
received the B.S. degree from Huazhong University of Science and Technology, Wuhan, China, in 2017, and he is currently working toward the Ph.D. degree in the Department of Automation, Tsinghua University, Beijing, China.
His research interests include machine learning, deep learning and computer vision.
\end{IEEEbiography}
\vspace{-2em}

\begin{IEEEbiography}[{\includegraphics[width=1in,height=1.25in,clip,keepaspectratio]{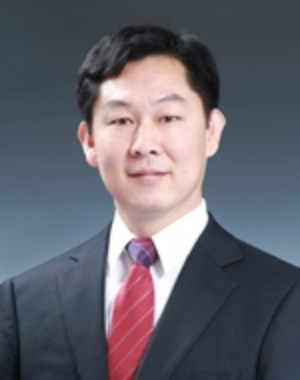}}]{Changshui Zhang}
(M'02--SM'15--F'18) received the B.S. degree in mathematics from Peking Univer- sity, Beijing, China, in 1986, and the Ph.D. degree from the Department of Automation, Tsinghua University, Beijing, in 1992.
He is currently a Professor with the Department of Automation, Tsinghua University. His current research interests include artificial intelligence, image processing, pattern recognition, machine learning, and evolutionary computation.
\end{IEEEbiography}
\vspace{-1em}


\vfill


\end{document}